# AI Predicts AGI: Leveraging AGI Forecasting and Peer Review to Explore LLMs' Complex Reasoning Capabilities

Fabrizio Davide (1); Pietro Torre (1); Leonardo Ercolani (2); Andrea Gaggioli (3, 4)

(1) ISTAT, Rome Italy

(2) Department of Advanced Computing Sciences, Maastricht University, Maastricht, The Netherlands

(3) Research Center in Communication Psychology (PSICOM), Università Cattolica del Sacro Cuore, Milan, Italy;

(4) IRCCS Istituto Auxologico Italiano, Milan, Italy

**Abstract**

We tasked 16 state-of-the-art large language models (LLMs) with estimating the likelihood of Artificial General Intelligence (AGI) emerging by 2030. To assess the quality of these forecasts, we implemented an automated peer review process (LLM-PR). The LLMs' estimates varied widely, ranging from 3% (Reka-Core) to 47.6% (GPT-4o), with a median of 12.5%. These estimates closely align with a recent expert survey that projected a 10% likelihood of AGI by 2027, underscoring the relevance of LLMs in forecasting complex, speculative scenarios. The LLM-PR process demonstrated strong reliability, evidenced by a high Intraclass Correlation Coefficient (ICC = 0.79), reflecting notable consistency in scoring across the models. Among the models, Pplx-70b-online emerged as the top performer, while Gemini-1.5-pro-api ranked the lowest. A cross-comparison with external benchmarks, such as LMSYS Chatbot Arena, revealed that LLM rankings remained consistent across different evaluation methods, suggesting that existing benchmarks may not encapsulate some of the skills relevant for AGI prediction. We further explored the use of weighting schemes based on external benchmarks, optimizing the alignment of LLMs' predictions with human expert forecasts. This analysis led to the development of a new, 'AGI benchmark' designed to highlight performance differences in AGI-related tasks. Our findings offer insights into LLMs' capabilities in speculative, interdisciplinary forecasting tasks and emphasize the growing need for innovative evaluation frameworks for assessing AI performance in complex, uncertain real-world scenarios.

## 1. Introduction

Large Language Models (LLMs) are a type of artificial intelligence system trained on vast amounts of text data to understand and generate human-like text. These models, which include systems like GPT (Generative Pre-trained Transformer) and BERT (Bidirectional Encoder Representations from Transformers), have demonstrated remarkable capabilities in tasks ranging from natural language understanding and generation to complex problem-solving and analysis. As these models continue to evolve, becoming increasingly sophisticated and multifaceted, the need for comprehensive evaluation methods has become paramount.

Traditional evaluation methods for LLMs often rely on task-specific benchmarks designed to assess performance in narrowly defined domains. Standardized tests in areas such as question answering, text summarization, or sentiment analysis provide important insights into specific functionalities. However, these tests often operate within confined parameters that may not reflect the open-ended, multifaceted nature of real-world cognitive challenges.

The limitations of traditional benchmarks become particularly apparent when attempting to evaluate LLMs' performance on tasks that require the integration of knowledge across multiple domains, abstract reasoning, and metacognitive abilities. Real-world problems often demand a synthesis of diverse information, the ability to reason under uncertainty, and the capacity for self-evaluation—facets that existing evaluation frameworks may not fully capture.

To address these limitations, we introduce an assessment methodology that combines two key tasks:

**1. An Artificial General Intelligence (AGI) forecasting task**: We tasked LLMs with predicting the probability of AGI occurring by 2030. We chose this task because it presents an open-ended challenge requiring the integration of knowledge across multiple domains such as computer science, cognitive science, philosophy, and futurism.

**2. A LLM peer review (LLM-PR) task**: This approach involves LLMs evaluating each other's forecasts, including their own, based on a set of predefined criteria. This method builds upon and extends previous work on using LLMs for evaluation of LLMs output, such as the "LLM as a Judge" approach introduced by Zheng et al., (Zheng et al., 2024). Given the complex and speculative nature of AGI forecasting, we included structured criteria to break down the LLM evaluation of AGI forecasts into specific components. We formulated the following research questions to guide our exploratory study:

1. How does the performance of LLMs compare to human experts in forecasting AGI development?
2. How do LLMs assess their own forecasts and those of other models? To what extent are these self and peer evaluations reliable and consistent?
3. Is there a correlation between an LLM's performance on traditional benchmarking tasks and the quality of its AGI forecasts and peer ratings?

LLMs AGI predictions ranged from 3% (Reka-Core) to 47.6% (gpt-4o), with a median of 12.5%, closely aligning with a recent expert survey on AGI timelines (Grace et al., 2024), which estimated a 10% aggregate probability of AGI by 2027. Key themes in the LLMs forecasts included the role of machine learning advances, hardware improvements, and interdisciplinary research, alongside concerns about ethical and regulatory barriers. The LLM-PR process revealed high consistency in scoring (ICC = 0.79), with pplx-70b-online scoring the highest and Gemini-1.5-pro-api the lowest. Interestingly, the use of traditional benchmarks, such as LMSYS Chatbot Arena, to score the forecasts revealed consistent rankings across various methods, indicating that these benchmarks may capture some skills relevant to AGI prediction. However, our analysis also showed that these benchmarks may not fully encompass the specific capabilities required for forecasting complex and speculative scenarios like AGI. To address this, we refined the weighting schemes and optimized them to achieve closer alignment with human expert evaluations. This process led to the development of a new 'AGI benchmark,' specifically designed to highlight performance variations among models in the context of AGI, offering a more tailored and precise evaluation framework.

The paper is organized as follows. We begin by discussing current challenges in evaluating LLMs and explaining our use of AGI forecasting as a proxy for assessing complex reasoning capabilities. In sections 3 and 4 we introduce the AGI forecasting task submitted to a panel of LLMs and analyze the outcomes. In sections 5 and 6 we detail our methodology for the LLM peer review (LLM-PR) process and discuss our findings. Finally, we provide a comparison of results with an expert survey results and introduce a new benchmark related to the AGI forecast. The discussion explores the implications of our results for LLM development and evaluation. We conclude by summarizing key insights and proposing future research directions in this area.

## 2. Background

### *2.1 Evolving challenges in LLMs evaluation*

LLMs such as GPT, BERT, and their successors, have revolutionized natural language processing by demonstrating unprecedented capabilities in generating, understanding, and interacting with human language across a wide range of contexts. These models are trained on massive datasets and leverage sophisticated architectures to mimic human-like text generation and comprehension. As these models have rapidly advanced in capabilities, traditional evaluation methods that rely on narrow, task-specific benchmarks have become increasingly inadequate for assessing their full spectrum of abilities (McIntosh et al., 2024). The current landscape of LLM evaluation is fragmented, with a proliferation of benchmarks that lack standardization and may not accurately reflect real-world application scenarios (Tikhonov & Yamshchikov, 2023). This creates challenges in comprehensively and fairly comparing different LLMs, especially as they approach or potentially surpass human-level performance on many tasks. Moreover, the rapid pace of LLM development has outstripped the evolution of evaluation methodologies, leading to a situation where benchmarks quickly become obsolete or fail to capture the capabilities of the latest models (McIntosh et al., 2024). Furthermore, as Tikhonov and Yamshchikov (2023) point out, since LLMs increasingly mimic human-like behaviors, traditional evaluation proxies such as the Turing test have become less reliable, emphasizing the need for more flexible, holistic, and interdisciplinary approaches to LLM evaluation that can keep pace with rapid advancements in the field and provide meaningful insights into these models' true capabilities and limitations. Such approaches should not only assess technical performance but also consider ethical implications, robustness, and the ability to generalize across diverse tasks and domains (McIntosh et al., 2024). To contribute to this open challenge, we designed two tasks: one focused on forecasting the emergence of AGI, requiring models to integrate interdisciplinary knowledge and address uncertainty and temporal complexity, thereby testing their capabilities beyond traditional benchmarks. Additionally, we implemented the LLM Peer Review task, where LLMs evaluate their own forecasts and those of other models based on a structured set of criteria. This dual-task approach allows us to assess both the predictive accuracy and the evaluative consistency of the models, providing a comprehensive evaluation of their capabilities in the context of AGI forecasting.

*2.2 AGI forecasting task*

AGI refers to AI systems capable of performing any intellectual task that humans can, with comparable or superior proficiency across a wide range of domains (Goertzel & Pennachin, 2006). Also termed Human-Level Machine Intelligence (HLMI) or Human-Level AI (HLAI) (e.g., Besold and Schmid 2016), AGI surpasses narrow AI, which excels at specific, predefined tasks but lacks the adaptability and generalization capabilities of human intelligence. The potential impact of AGI on society is profound and multifaceted. In science and technology, AGI could accelerate research and innovation, potentially leading to breakthroughs in areas such as medicine, clean energy, and space exploration. In economics, AGI could dramatically increase productivity and economic growth, potentially reshaping labor markets and economic structures (Hanson, 2016). However, the development of AGI also raises significant ethical and existential concerns, including the potential for rapid, uncontrolled self-improvement leading to an intelligence explosion, as well as issues of AI alignment and control (Bostrom, 2014). Given AGI far-reaching implications, forecasting its development has become a subject of significant interest and debate. Several notable studies have attempted to gauge expert opinion on AGI timelines. Baum et al. (2011) surveyed participants at an AGI conference, finding that a majority expected human-level AGI to be achieved by 2050. The study revealed a dichotomy between "AGI optimists" and "AGI pessimists," with optimists generally expecting AGI within a few decades and pessimists projecting much longer timelines or expressing skepticism about AGI's feasibility. Müller and Bostrom (2016) conducted a global survey of AI experts, finding a wide range of opinions but a general trend towards expecting AGI within the 21st century. Their study also explored experts' views on the potential consequences of AGI development, including both positive and negative outcomes. Grace et al. (2018) surveyed a broader group of machine learning researchers, revealing a median estimate of 45 years until the achievement of high-level machine intelligence. This study also explored researchers' beliefs about the potential impacts of AGI, including economic, social, and existential risks. Zhang et al. (2022) carried out a comprehensive survey of AI and machine learning (ML) researchers regarding their views on AI advancements indicates that, on average, the respondents estimated a 50% probability of achieving human-level machine intelligence by 2060. More recently, a survey of 2,778 AI researchers provided a median forecast with a 50% probability that AI systems would achieve significant milestones by 2028 and that unaided machines would surpass human performance in all tasks by 2047. The same study estimated a 10% aggregate probability that AGI could be achieved by 2027 (Grace et al., 2024).

Recent works have examined the use of LLMs in specialized forecasting tasks. For instance, Chang et al. (2024) and Gruver et al. (2024) have shown that LLMs can predict future values in time series data with performance comparable to traditional statistical methods. Similarly, Schoenegger et al. (2023) and Halawi et al. (2024) have explored LLMs' ability to forecast real-world events, demonstrating that in some scenarios, LLMs can match or even surpass human crowd performance. However, to the best of our knowledge, no study has yet explored the use of LLMs in forecasting AGI development. AGI forecasting requires the integration of knowledge from various fields, including computer science, cognitive science, neuroscience, and philosophy, allowing to test LLMs' ability to synthesize information across diverse domains. It involves understanding and extrapolating technological trends over extended periods, challenging LLMs' capacity for long-term and temporal reasoning. Also, the task inherently involves dealing with high levels of uncertainty, testing LLMs' ability to reason probabilistically and qualify their predictions. Crucially, unlike many traditional benchmark tasks, there's no definitive "correct" answer in AGI forecasting.

*2.3 LLMs Peer Review task*

Recent research has explored various approaches to leveraging LLMs for self-evaluation. These methods aim to provide scalable, cost-effective alternatives to human evaluation while maintaining high levels of accuracy and insight. Liu et al. (2023) developed G-EVAL, a framework that uses LLMs to assess the quality of generated texts through a form-filling paradigm. The process involves providing the LLM with a task introduction and evaluation criteria, after which the LLM generates a chain-of-thought (CoT) detailing the evaluation steps. The LLM then uses this CoT to evaluate the text outputs in a structured manner. G-EVAL's approach allows for more fine-grained and explainable evaluations, as the LLM not only provides scores but also rationales for its judgments. The authors found that G-EVAL, particularly when using GPT-4, achieved higher correlations with human judgments compared to previous methods, especially for open-ended tasks like dialogue generation. GPTScore (Fu et al., 2023) leverages the capabilities of LMS to assess the quality of generated text. This approach employs models like GPT-3 to assign higher probabilities to high-quality content through multidimensional evaluation prompted by multiple queries.

Dubois et al. (2024) introduced AlpacaEval, a benchmark specifically designed for evaluating instruction-following capabilities of chat models. AlpacaEval operates on a fixed set of instructions chosen to represent typical user interactions. The evaluation process involves both a baseline model and the evaluated model producing responses to the instructions, after which a GPT-4-based rater compares the responses head-to-head. A win rate is computed as the probability that the rater prefers the evaluated model's output. To address potential biases, particularly length bias, the authors developed AlpacaEval-LC (Length-Controlled). This version uses a regression-based approach to estimate what the preference would be if the outputs of all models had the same length as the baseline. AlpacaEval-LC showed improved correlation with human judgments and increased robustness against output verbosity.

Some researchers have also proposed approaches that use multiple LLMs as evaluators. ChatEval (Chan et al., 2023), introduced a multi-agent evaluation framework that simulates the human evaluative process through a multi-agent debate to enhance the automated assessment of text generation quality. Similarly, PRE (Chu et al., 2024) and PRD (Li et al., 2023) have advocated for the use of LLMs as evaluators, combining multiple evaluation outcomes for the automated assessment of other LLMs' performance. Recently, Ning et al. (2024) proposed PiCO (Peer review in LLMs based on Consistency Optimization), an unsupervised approach for evaluating LLMs without human feedback. Their method utilizes a peer-review mechanism where LLMs evaluate each other's responses to open-ended questions. PiCO assigns a learnable capability weight to each LLM and optimizes these weights to maximize consistency between an LLM's capability and its evaluation scores. In contrast with supervised methods like PRE, which uses human feedback throughout the evaluation process, PiCO's approach aims to create a ranking of LLMs that aligns with human preferences, while operating in a fully unsupervised manner. Zheng et al. (2023) introduced the "LLM-as-a-Judge" method, which utilizes advanced LLMs like GPT-4 to evaluate the outputs of other models. This approach employs either pairwise comparisons or single-answer grading, where the LLM judge is presented with a question and two answers (or a single answer) and tasked with determining which is better or assigning a score. The authors propose three variations of this method: pairwise comparison, single answer grading, and reference-guided grading. In pairwise comparison, the LLM judge decides which of two responses is better or declares a tie. Single answer grading involves the LLM judge directly assigning a score to a single answer. Reference-guided grading provides the LLM judge with a reference solution, particularly useful for tasks like math problems. The study demonstrated that LLM judges, particularly GPT-4, could achieve over 80% agreement with human evaluations, matching the level of agreement between humans. This high level of correspondence suggests that LLM-as-a-Judge could serve as a reliable proxy for human evaluation in many scenarios.

The LLM Peer Review (LLM-PR) method that we developed in this study builds upon these foundations while introducing two novel features. In LLM-PR, each LLM evaluates not only others but also itself, potentially offering a more comprehensive and reflective perspective on model quality. Furthermore, our method extends the concept of LLM-based evaluation by incorporating a structured set of criteria measured on a Likert-type scale for assessments. We suggest that this approach may allow for a more granular and multifaceted evaluation compared to binary or holistic judgments. Additionally, by having each model serve as both subject and rater, LLM-PR may provide insights into the "metacognitive" capabilities of LLMs - their ability to critically assess their own performance. Metacognition, often described as "thinking about thinking," involves processes that monitor, regulate, and enhance cognitive functions. While in the context of LLMs these processes do not equate to true human-like metacognition, they represent an important step towards more sophisticated AI systems capable of self-evaluation and improvement. For example, Wang and Zhao (2024) introduced Metacognitive Prompting (MP) to enhance LLMs' understanding abilities in natural language understanding tasks. Their method guides LLMs through structured, self-aware evaluations, which model human introspective reasoning.

PiCO proposes an optimization procedure based on the assignment of confidence weights that will be learned during the review process and the alignment with human evaluations. To this end ad hoc metrics are used that are like the normalized Kendall distance we adopted. Finally, PiCO advances a consistency assumption that results not necessary in the ranking approach we deploy to align the unsupervised review results with human evaluations. This introduces a significant methodological variation and the need for a standardized approach that ensures comparability and reliability in LLM evaluations. We address this discrepancy and highlight the implications of using different evaluation criteria, while proposing adjustments to improve homogeneity and consistency.

## 3. AGI forecasting Task

### 3.1 LLMs

We selected 16 leading LLMs for this study, as listed in Table 1. The selection was based on the LMSYS Chatbot Arena ranking, an open and recognized platform for evaluating LLMs. LMSYS Chatbot Arena uses a crowdsourced evaluation method that has collected over 1,000,000 pairwise comparisons made by humans. The ranking is generated using the Bradley-Terry model and displayed on the Elo scale, which are well-established methods for comparative performance evaluation. The decision to limit to the top 16 Arena models (as updated on 12th July 2024) allowed us to conduct an in-depth analysis of each model, including peer-to-peer evaluation, while maintaining the feasibility of the study. Noteworthy, this selection includes both publicly accessible models and non-public models. This diversity allows us to explore how different architectures, sizes, and development philosophies influence the predictive and evaluative capabilities of LLMs.

**Table 1.** LLMs included in the study (PP: proprietary model; NP: non-proprietary model). The exact name and version of each LLM is reported in Appendix.

| ***LLM*** | ***PP/NP*** | ***Architecture*** | **Arena score (as per 2024-07- 17)** |
|---|---|---|---|
| *gpt-4o-2024-05-13* | *PP* | *Transformer* | 1282 |
| *claude-3-5-sonnet-20240620* | *PP* | *Transformer* | 1272 |
| *gemini-1.5-pro-api-0514* | *PP* | *Transformer* | 1267 |
| *Yi-Large-preview* | *PP* | *Transformer* | 1241 |
| *GLM-4-0520* | *NP* | *Transformer* | 1216 |
| *Llama-3-70b-Instruct* | *PP* | *Transformer* | 1207 |
| *Reka-Core-20240501* | *NP* | *Transformer* | 1207 |
| *Command-R+* | *PP* | *BERT-like* | 1200 |
| *Qwen2-72B-Instruct* | *NP* | *BERT-like* | 1190 |
| *DeepSeek-Coder-V2-Instruct* | *NP* | *BERT-like* | 1188 |
| *Mistral-Large-2402* | *NP* | *BERT-like* | 1179 |
| *Mixtral-8x22b-Instruct* | *PP* | *BERT-like* | 1157 |
| *Phi-3-Medium-4k-Instruct* | *NP* | *BERT-like* | 1146 |
| *Gemma-2-27B-it* | *NP* | *Others* | 1123 |
| *DBRX-Instruct-Preview* | *NP* | *Others* | 1103 |
| *pplx-70b-online* | *PP* | *Others* | 1078 |

3.2 AGI Forecasting task: procedure

Each LLM was presented with a detailed forecasting prompt asking them to estimate the likelihood of AGI occurring by late 2030. The prompt (see 11.1.1) included:

- a definition of AGI:

"Artificial General Intelligence (AGI), also known as Strong AI or Full AI, refers to a type of artificial intelligence that can understand, learn, and apply intelligence across a wide range of tasks at a level comparable to human beings."

- specific conditions for considering AGI achieved:

a) An AI system wins a journalism prize using a human pen name, with its work submitted and published without any editing or intervention by humans.

b) An AI system analyzes medical data on a specific type of cancer, collaborates with human researchers unaware they are interacting with an AI, and ultimately discovers a novel and unexpected treatment.

c) An AI agent autonomously manages a multinational corporation for a full fiscal year, making strategic decisions, conducting negotiations, and adapting to market changes without human intervention. The company achieves record profits and significantly outperforms industry benchmarks, while also implementing innovative sustainability practices that were not part of its original programming.

- a base rate of 1% for the AGI event occurring by late 2030.

Instructions to provide:

● A rationale for the estimation.

● An approach to forecasting.

● A likelihood estimation based on a mathematical or statistical model.

The prompt was set with a temperature of 1, Top P of 1, and max output tokens of 2000. These parameters were selected to allow for maximum diversity in token selection, enabling exploration of a wide range of scenarios. The 2000 token limit provides sufficient space for detailed reasoning and comprehensive responses without excessive verbosity.

## 4. Analysis of LLMs forecasts

To analyze the forecasts generated by the 16 LLMs, we performed a qualitative analysis of the text to capture key themes and patterns in the LLMs' reasoning. First, the codes for analysis were defined and applied to each LLM response. Following, a thematic analysis was conducted to identify overarching themes and patterns across the LLM responses. The analysis was performed using the software MAXQDA 2020.

### *4.1 Qualitative analysis of LLMs forecasts*

We first categorized the LLMs forecasts based on the probability assigned to AGI development by late 2030 (Table 2). The distribution of predictions shows that the majority of models (13 out of 16, or 81.2%) forecast a probability lower than 30% for an AGI event by 2030, with only 3 models (18.7%) being optimistic with predictions above 30%. Among the optimistic models, pplx-70b-online is the most confident with a 47% probability, closely followed by gpt-4o-2024-05-13 at 45%, and Yi-Large-preview at 38%. In the moderate range, 6 models (37.5% of the total) predict a probability between 10% and 30%, with estimates in this group varying from 12% to 15%. The pessimistic category, which includes most models (7 out of 16, or 43.7%), forecasts a probability below 10%, with estimates ranging from 3% to 8%. The overall trend leans towards more conservative predictions, with most models anticipating a low probability of an AGI event by 2030. However, there is a notable variation in estimates, spanning from 3% to 47%, indicating a high degree of uncertainty or disagreement among the models.

**Table 2.** Probability assigned by LLMs to the development of AGI by late 2030.

| **Probability prediction for an AGI event by 2030** | **Frequen -cy** | **Percen-tage** | **LLMs** |
|---|---|---|---|
| Optimistic (>30%) | 3 | 18.75% | pplx-70b-online (47%), gpt-4o (45%) Yi-Large-preview (38%) |
| Moderate (10-30%) | 6 | 37.5% | Qwen2-72B-Instruct (15%) Mixtral-8x22b-Instruct-v0.1 (15%) Mistral-Large-2402 (12%) Llama-3-70b-Instruct (15%) gemini-1.5-pro-api-0514 (12,5%) Command-R+ (15%) |
| Pessimistic (<10%) | 7 | 43.75% | Reka-Core (3%) Phi-3-Medium-4k-Instruct (6,3%) GLM-4-0520 (8%) Gemma-2-27B-it (5%) DeepSeek-Coder-V2-Instruct (5%) DBRX-Instruct-Preview (3,5%) claude-3-5-sonnet (5,8%) |
| Total | 16 | 100% | |

Most LLMs employed historical comparisons to contextualize AGI development by drawing parallels with previous technological milestones, to offer a reference point for understanding the complexities and uncertainties associated with developing AGI. For example, the development of the Internet was referenced by four LLMs as an example of a transformative technology that emerged over a few decades. Three LLMs cited the progress of narrow AI and machine learning to show the current state and trajectory of AI development. The advent of personal computers and smartphones were also mentioned by two LLMs as examples of technologies that significantly changed society. Other notable comparisons included the development of nuclear energy, cited by four LLMs, the invention of the microprocessor, the development of commercial flight, and the sequencing of the human genome. Some LLMs also referenced specific AI milestones to highlight the field's progression. For example, Mixtral-8x22b-Instruct-v0.1 mentioned Deep Blue (1997), Watson (2011), AlphaGo (2016), and GPT-3 (2020) as indicators of accelerating progress in AI.

All LLMs discussed factors that could accelerate or hinder AGI development, to highlight the complexity of predicting AGI's timeline and the multitude of factors that can influence its progress. Among the most frequently cited inciting events, advances in machine learning algorithms and breakthroughs in computational power. Advances in hardware were also noted, as well as the developments in cognitive neuroscience. Significant investments in AI research were cited as crucial factors that can accelerate progress towards AGI. Increased funding and resources dedicated to AI research was mentioned as a factor that can lead to more research initiatives, talent acquisition, and resource availability. On the blocking side, ethical and regulatory constraints were the most frequently mentioned, underscoring the potential impact of safety, fairness, and societal concerns in slowing down or halting AGI development. Limitations in current research methodologies were pointed out by several LLMs, indicating that current approaches in AI research might not be sufficient to achieve AGI, requiring new paradigms and innovative solutions. Unforeseen technical stumbling blocks were acknowledged by 8 LLMs, highlighting the unpredictable nature of scientific and technological challenges as factors that could introduce significant delays in achieving AGI. Several LLMs cited specific trends and trajectories to define the context for predicting AGI development. For example, 4 LLMs referenced Moore's Law (which predicts the doubling of transistors on integrated circuits approximately every two years) to describe the context of technological evolution eventually leading to AGI.

In describing the forecasting approach, the majority of LLMs (10/16) recognized the development of AGI as a complex and ambitious goal with significant uncertainties and potential roadblocks. Frequently cited factors contributing to the uncertainty include the difficulty in predicting the pace of technological progress, potential regulatory or societal barriers, and the need for fundamental advances in our understanding of human intelligence and cognition. For instance, one LLM noted, "Previous forecasts for AGI have varied widely, with some suggesting feasibility as early as 2030 and others predicting much later dates or even questioning the possibility altogether" (Reka-Core-20240501). Another stated, "Given the inherent uncertainty in predicting such a complex phenomenon, I must stress that this estimation is based on the assumption that the current base rate remains constant over time, with no major inciting or blocking events radically shifting the overall progress of AGI" (Phi-3-Medium-4k-Instruct). These examples underscore the cautious approach taken by LLMs in their predictions, highlighting the significant challenges and uncertainties involved in developing AGI. Furthermore, about half of LLMs (7/16) consider recent analysis and predictions for AGI (such as Ray Kurzweil's prediction of 2029 for AGI), varying predictions underscore the complexity and transformative potential of AGI. For example, one LLM stated, "Recent predictions for AGI have ranged from approximately 2030-2045 with base rates around 5-10%. Notable past forecasts such as Ray Kurzweil's predictions which suggest a timeline of 2029 for AGI can provide an insightful context" (Phi-3-Medium-4k-Instruct). Another LLM mentioned, "Examining expert predictions, there is a wide range of opinions reflecting the topic's inherent uncertainty" (Mistral-Large-2402).

Some LLMs stressed the importance of interdisciplinary considerations in predicting AGI development. For example, Yi-Large-preview emphasized the multifaceted nature of AGI development, involving advances in machine learning, hardware capabilities, energy efficiency, and interdisciplinary collaborations. GLM-4-0520 highlighted the need for technological breakthroughs, algorithmic innovations, and new conceptual frameworks, with fields like neuroscience, psychology, and philosophy influencing AGI's trajectory.

Table 3 lists the mathematical models and equations used by LLMs to estimate the probability of AGI development by late 2030. The use of Bayesian approaches is prevalent among the LLMs (10/16), highlighting the importance given by most LLMs to update prior beliefs with new evidence in forecasting AGI development. The probability estimates from these models range from 2-4% for Reka-Core-20240501 to 47.62% for pplx-70b-online, indicating a broad spectrum of confidence levels. Statistical and regression models were applied by Phi-3-Medium-4k-Instruct, which used a Poisson regression model considering the time variable, and by Mistral-Large-2402, which used a logistic growth model, leading to a 12% probability. GPT-4o-2024-05-13 implemented a statistical growth model with time-dependent variables.

**Table 3.** Mathematical models and equations used by LLMs to estimate the probability of AGI development by 2030.

| LLM Name | Model/Equation Used | Probability Estimate by 2030 (%) |
|---|---|---|
| Yi-Large-preview | Bayesian approach updating the base rate probability (1%) | 38 |
| Reka-Core-2 | Bayesian approach updating the base rate probability (1%) | 3 |
| Qwen2-72B-Instruct | Bayesian approach updating the base rate probability (1%) | 15 |
| Mixtral-8x22b-Instruct-v0.1 | Bayesian approach updating the base rate probability (1%) | 15 |
| GLM-4-0520 | Bayesian approach updating the base rate probability (1%) | 8 |
| DeepSeek-Coder-V2-Instruct | Bayesian approach updating the base rate probability (1%) | 5 |
| pplx-70b-online | Bayesian approach updating the base rate probability (1%) | 47,6 |
| DBRX-Instruct-Preview | Bayesian approach updating the base rate probability (1%) | 3,5 |
| Llama-3-70b-Instruct | Bayesian approach using modified log-normal distribution model | 15 |
| Phi-3-Medium-4k-Instruct | Poisson regression model considering the time variable | 6,3 |
| Mistral-Large-2402 | Logistic growth model | 12 |
| gemini-1.5-pro-api-0514 | Modified logistic function incorporating time and accounting for potential acceleration | 12,5 |
| claude-3-5-sonnet | Modified Gompertz function for technological adoption and breakthrough probabilities | 5.8 |
| gpt-4o-2024-05-13 | Statistical growth model with time-dependent variables | 45 |
| Gemma-2-27B-it | Bayesian approach updating the base rate probability (1%) | 5 |
| Command-R+ | Model/equation not specified | 15 |
| | Mean | 15,73 |
| | Standard deviation | 14,11 |
| | Median | 12,25 |

Gemini-1.5-pro-api-0514 used a modified logistic function incorporating time and potential acceleration. Claude-3-5-sonnet utilized a modified Gompertz function for technological adoption and breakthrough probabilities. One model, Gemma-2-27B-it, did not specify a model or equation but provided a 5% probability estimate.

### *4.2 Comparison of LLM-based and human experts AGI forecasts*

To compare the predictions generated by LLMs with human expert forecasts, we used the results from the survey "Thousands of AI Authors on the Future of AI" by Grace et al., (2024). The survey, conducted in 2023, involved 2,778 researchers who had published in six top-tier AI venues, which provides a fair representation of expertise within the AI research community. The survey defined High-Level Machine Intelligence (HLMI) and asked participants to predict when it would be feasible, assuming continued scientific progress. Of the total participants, 1,714 answered the HLMI question. The survey employed both fixed-year and fixed-probability question formats to reduce potential framing biases. Each participant provided three year-probability pairs, which were used to fit a gamma distribution. These individual distributions were then aggregated by calculating the mean across all participants. This approach yielded a median probability equal to 10% of achieving high-level machine intelligence (HLMI) by 2027. The alignment in AGI probability estimates between LLM predictions of AGI by 2030 and those of human experts of AGI by 2027 (respectively, 12.25% vs. 10%) confirms previous observations that LLMs are not only capable of performing forecasting tasks but also able to produce results comparable to human predictions (Schoenegger et al., 2023; Halawi et al., 2024). However, in the context of this study, the reference to human expert forecasts is not intended to assess how closely LLM performance in the forecasting task matches or diverges from human performance, as would be the case in “standard” benchmarking tasks (e.g., coding challenges, math problems, real-world science questions). In fact, such a comparison would not even be appropriate, since the prompt given to the LLMs did not coincide with the instructions provided to the experts in the Grace study. Instead, the reference to human experts serves to provide context and a useful point of comparison for understanding the reasoning and justifications produced by the LLMs.

## 5. LLM Peer-review task

### *5.1 Peer-evaluation procedure*

To further evaluate the forecasters' output, we considered using a panel of human experts, such as futurologists and professional forecasters. However, we opted to have the LLMs evaluate themselves. This approach not only allowed us to assess the models' predictive abilities but also provided insight into how they evaluate one another and their self-assessment skills. Crucially, this approach enables a comparative analysis of the LLMs' ability to both *generate* and *assess* forecasts, highlighting strengths and weaknesses in different aspects of their reasoning.

Accordingly, each LLM (i.e., rater or judge) was tasked with evaluating the responses of the other LLMs (i.e., forecasters) based on nine specific criteria (listed in Table 4). These criteria were carefully designed to measure the quality, depth, and rigor of the LLMs' forecasts, including the structure of their reasoning, the richness of the provided context, and the appropriateness of the statistical models applied.

**Table 4.** Evaluation criteria used by LLMs raters to assess LLMs forecasts**.**

| Code | Forecast evaluation criterion | Description | Aspect evaluated | Scoring |
|---|---|---|---|---|
| C1 | Well-structured and thoroughly documented rationale for the likelihood estimation | Evaluates the clarity, logic, and explanation of the LLM's reasoning process. A strong rationale is fundamental for understanding and critically evaluating the prediction. | Qualitative reasoning | 1-5 Likert scale (1 completely disagree; 5 completely agree) |
| C2 | Non-trivial comparisons to analogous or similar events and technological advancements. | Assesses the LLM's ability to draw relevant parallels and learn from historical precedents, demonstrating a deeper understanding of technological progress patterns. | Qualitative reasoning | 1-5 Likert scale (1 completely disagree; 5 completely agree) |
| C3 | Rich context for the AGI event, including potential catalysts and obstacles. | Evaluates the LLM's ability to consider a wide range of factors influencing AGI development, crucial for informed predictions about complex technological advancements. | Qualitative reasoning | 1-5 Likert scale (1 completely disagree; 5 completely agree) |
| C4 | Thorough discussion of the provided base rate. | Assesses the LLM's understanding of probabilistic reasoning and its ability to incorporate given information into its forecast, a key skill in accurate forecasting. | Use of historical data and expert knowledge | 1-5 Likert scale (1 completely disagree; 5 completely agree) |
| C5 | Reporting on relevant past events and other pertinent forecasts. | Evaluates the LLM's ability to research and incorporate historical data and expert opinions, testing its capacity to synthesize information from various sources. | Use of historical data and expert knowledge | 1-5 Likert scale (1 completely disagree; 5 completely agree) |
| C6 | Comprehensive examination of unexpected breakthroughs. | Assesses the LLM's ability to consider low-probability, high-impact events that could significantly alter the AGI development timeline, crucial for comprehensive forecasting of transformative technologies. | Consideration of uncertainty and extreme events | 1-5 Likert scale (1 completely disagree; 5 completely agree) |
| C7 | Use of an appropriate and sufficiently complex statistical model. | Evaluates the LLM's ability to apply quantitative methods to forecasting, testing whether it can provide a structured, mathematical approach to prediction. | Quantitative modeling skills | 1-5 Likert scale (1 completely disagree; 5 completely agree) |
| C8 | Clear description of model parameters consistent with given conditions and analysis. | Ensures the LLM's forecasting process is transparent and replicable, testing its ability to explain complex concepts clearly. | Quantitative modeling skills | 1-5 Likert scale (1 completely disagree; 5 completely agree) |
| C9 | Demonstrably fair and reasonable evaluation of model parameters. | Assesses the LLM's ability to make balanced judgments and avoid biases in its forecasting process, ensuring the final prediction is based on well-reasoned parameter estimates. | Quantitative modeling skills | 1-5 Likert scale (1 completely disagree; 5 completely agree) |

### *5.2 Scoring model*

We use a single-point scoring model, where each rater evaluates the quality of a forecast independently, without direct comparison to other forecasts (Verga et al., 2024). The evaluation prompt (see 11.1.2) provides clear instructions on how the grading should be conducted, defining the characteristics of a good or poor response. Thus, ratings are based solely on the rater's judgment of what constitutes a high-quality forecast. Therefore, the j-th rater independently scores the i-th forecast, after the k-th criterion, with a score $s_{ij}^{(Ck)} \in \{1,2 \ldots 5\}$. Those individual scores are then pooled together forming the matrix $S^{(C)} = \left[S^{(C1)} \middle| S^{(C2)} \middle| \ldots S^{(C9)}\right] \in R^{16\times 153}$. The final score $s_i$ of the i-th forecaster after the panel voting needs a counting function $f$ to be computed: $s_i = f\left(s_{ij}^{(Ck)}, j =\right.$

$1..16, k = 1..9$). Here we will often assume to reduce $S^{(C)}$ to $S$ averaging over the criteria:

$$S = \frac{1}{9}\sum_{k=1}^{9} \quad S^{(Ck)} \in R^{16\times 16}$$

where $s_{ij}$ as the generic element of $S$ represents the average score across the criteria given by the j-th rater to the i-th forecaster. A simple example of the counting function is a weighted sum of $s_{ij}$ (computed after averaging across the criteria), resulting in a forecaster's score as in equ. 1:

$$s_i = \sum_{j=1}^{16} \quad w_j s_{ij} \qquad (1)$$

with $w$ a suitable L1-normalized weight vector, whose j-th component $w_j$ represents the weight assigned to the j-th rater. We will often use $w_j$ constant with $j$ and call the resulting forecasters scores $s_i$ as "uniformly weighted scores", or simply "uniform scores".

*5.3 Results of the peer review*

Table 5 presents the scores assigned by the raters (listed horizontally) to the forecasters (listed vertically), averaged across the criteria, averages over the LLMs ensemble and the standard deviation of each rater's scores. On average, DBRX-Instruct-Preview (X15) was the most generous, assigning an average score of 4.8. In contrast, Gemini-1.5-Pro-API-0514 (X3) was the most critical, with an average score of 2.7. Further, Gemini 1.5 Pro exhibited the highest coefficient of variation in given scores, indicating significant differences in its evaluations. The standard deviation of the scores assumes the maximum (3,8) for the rater Command-R+ and the minimum (0,10) for the rater Gemma-2.

Fig. 1 reports the studentized residuals of the scores (i.e. the dimensionless ratio resulting from the division of a residual by the sample estimate of its standard deviation, as the mean and standard deviation are estimated per rater on the LLM ensemble). The distribution of the studentized residuals is consistent between the raters, as the ICC analysis in section 6.1 will explain in depth, meaning that all the raters contribute significantly to the final scoring and ranking.

Table 6 presents the evaluation scores assigned to each LLM for its AGI forecasts for each of the nine criteria, after averaging across all raters. This is computed as: $s_i^{(Ck)} = \frac{1}{16}\sum_{j=1}^{16} \quad s_{ij}^{(Ck)} \in R^{16\times 9}$.Overall, the scores are high, with a grand mean score of 4.207. Scores range from a low of 3.500 (Gemini-1.5-pro-api on criterion 5) to a high of 4.938 (pplx-70b-online on criterion 3). "Rich context for the AGI event" (Criterion 3) received the highest average score of 4.52, indicating that according to LLM raters, most forecasts excelled in providing comprehensive contextual information. In contrast, "Reporting on relevant past events and other pertinent forecasts" (Criterion 5) had the lowest average score of 3.98, suggesting it was a common area of weakness. The standard deviation of the criterium score assumes the maximum (0,26) with Criterium 6 and the minimum (0,19) with Criterium 1.

Fig. 2 presents the studentized residuals of the same scores as in Table 6. The distribution of the studentized residuals is consistent between the criteria, as the ICC analysis in section 6.1 will show, meaning that all the criteria contribute significantly to the final scoring and ranking.

Fig. 3 displays the rankings determined by each rater, plus the final ranking of forecasters, based on the uniform weighting of the raters scores, shown on the far right. Let us focus of the Top3 forecasters and their ranks along the raters: pplx-70b is ranked as a Top3 by 9 (over 16) raters, and often falls to the lowest ranks (11, 12, 13, 15); Qwen2 is ranked Top3 by 4 raters, keeping the 9-th rank with three raters, even resulting 9 and 11; Llama-3-70b is ranked Top3 by 3 raters, keeping the 6-th rank with three raters, and resulting 9 and 10. This means that the raters have a significant diversity of evaluations, and the weight that we give to each of them can affect the final ranking.

Fig. 4 shows that pplx-70b is ranked as a Top3 by 8 criteria over 9 with just one drop to 10; Qwen2 is ranked Top3 only by 5 criteria, Llama-3-70b is ranked Top3 only by 3 criteria.

**Table 5.** The evaluation scores averaged across all criteria: cell(j,k) of the table is computed as $s_{ij} = \frac{1}{9}\sum_{k=1}^{9} \quad s_{ij}^{(Ck)} \in R^{16\times 9}$ . The last columns reports the uniformly weighted average of the forecaster scores: $s_i = \frac{1}{16}\sum_{j=1}^{16} \quad s_{ij}$ and the ranking index in the forecaster ensemble. (U.S.: Uniform Score, R.I.: Ranking Index).

| | **X1** | **X2** | **X3** | **X4** | **X5** | **X6** | **X7** | **X8** | **X9** | **X10** | **X11** | **X12** | **X13** | **X14** | **X15** | **X16** | **US** | **RI** |
|---|---|---|---|---|---|---|---|---|---|---|---|---|---|---|---|---|---|---|
| **gpt-4o** | 3,89 | 4,22 | 3,11 | 3,89 | 3,11 | 4,00 | 4,56 | 4,44 | 4,33 | 4,89 | 4,56 | 4,56 | 4,44 | 4,56 | 4,56 | 4,33 | 4,22 | 7 |
| **claude-3-5-sonnet** | 4,78 | 4,56 | 3,22 | 3,78 | 4,00 | 4,00 | 4,89 | 4,56 | 4,00 | 4,67 | 4,22 | 4,44 | 4,44 | 4,56 | 4,44 | 4,44 | 4,31 | 4 |
| **gemini-1_5-pro** | 3,78 | 3,67 | 2,44 | 3,67 | 3,00 | 3,78 | 4,56 | 4,44 | 4,00 | 4,78 | 4,00 | 4,33 | 4,44 | 4,11 | 4,33 | 4,22 | 3,97 | 16 |
| **Yi-Large** | 4,67 | 4,67 | 3,11 | 3,89 | 3,78 | 4,33 | 4,00 | 4,56 | 4,44 | 4,78 | 4,67 | 4,00 | 4,78 | 4,00 | 4,78 | 4,56 | 4,31 | 4 |
| **GLM-4** | 3,78 | 3,89 | 2,56 | 3,44 | 2,89 | 3,89 | 3,89 | 4,56 | 4,11 | 4,78 | 4,11 | 3,89 | 4,33 | 5,00 | 4,78 | 4,67 | 4,16 | 11 |
| **Llama-3-70b** | 4,11 | 4,11 | 2,44 | 4,00 | 3,00 | 4,22 | 4,89 | 4,56 | 4,11 | 4,78 | 4,11 | 4,67 | 4,33 | 3,89 | 4,89 | 4,44 | 4,33 | 3 |
| **Reka-Core** | 4,11 | 4,67 | 2,67 | 3,89 | 3,44 | 4,11 | 5,00 | 4,56 | 4,11 | 4,78 | 4,22 | 4,22 | 5,00 | 4,56 | 5,00 | 4,89 | 4,01 | 15 |
| **Command-R+** | 4,00 | 4,00 | 2,44 | 3,67 | 3,00 | 4,00 | 3,89 | 4,56 | 4,11 | 4,78 | 4,11 | 4,11 | 4,67 | 3,44 | 4,89 | 4,56 | 4,17 | 10 |
| **Qwen2** | 4,00 | 3,89 | 2,56 | 3,89 | 3,44 | 4,44 | 4,67 | 4,56 | 4,11 | 4,78 | 4,11 | 4,56 | 4,33 | 3,67 | 4,78 | 4,89 | 4,38 | 2 |
| **DeepSeek-Coder** | 4,44 | 4,56 | 3,11 | 3,44 | 3,67 | 4,44 | 5,00 | 4,56 | 4,11 | 4,78 | 4,11 | 4,56 | 4,89 | 4,33 | 5,00 | 5,00 | 4,11 | 13 |
| **Mistral-Large** | 4,22 | 3,89 | 2,44 | 3,44 | 3,22 | 4,33 | 3,89 | 4,44 | 4,00 | 4,78 | 4,11 | 4,33 | 4,56 | 4,22 | 4,89 | 5,00 | 4,19 | 8 |
| **Mixtral-8x22b** | 4,44 | 4,33 | 2,44 | 3,44 | 3,33 | 4,22 | 4,33 | 4,56 | 4,11 | 4,78 | 4,11 | 4,22 | 4,78 | 4,00 | 4,89 | 5,00 | 4,19 | 8 |
| **Phi-3-Medium** | 4,33 | 3,89 | 2,56 | 3,67 | 3,33 | 4,11 | 4,67 | 4,56 | 4,11 | 4,78 | 4,11 | 4,00 | 4,89 | 4,33 | 4,89 | 4,89 | 4,16 | 11 |
| **Gemma-2** | 4,11 | 4,00 | 2,44 | 3,56 | 3,33 | 4,11 | 5,00 | 4,56 | 4,11 | 4,78 | 4,11 | 3,67 | 4,56 | 4,56 | 4,89 | 4,78 | 4,03 | 14 |
| **DBRX-Instruct** | 4,11 | 3,89 | 2,56 | 3,67 | 3,33 | 4,11 | 5,00 | 4,56 | 4,11 | 4,78 | 4,11 | 4,67 | 5,00 | 4,44 | 5,00 | 4,89 | 4,26 | 6 |
| **pplx-70b** | 4,89 | 4,67 | 3,11 | 4,56 | 4,00 | 4,33 | 4,11 | 4,89 | 4,67 | 4,78 | 4,67 | 4,78 | 4,78 | 4,11 | 4,78 | 5,00 | 4,51 | 1 |
| ***Average*** | ***4,23*** | ***4,18*** | ***2,70*** | ***3,74*** | ***3,37*** | ***4,15*** | ***4,52*** | ***4,56*** | ***4,16*** | ***4,78*** | ***4,22*** | ***4,31*** | ***4,64*** | ***4,24*** | ***4,80*** | ***4,72*** | ***4,21*** | **-** |
| ***Std dev.*** | ***0,33*** | ***0,33*** | ***0,30*** | ***0,28*** | ***0,34*** | ***0,19*** | ***0,43*** | ***0,10*** | ***0,17*** | ***0,04*** | ***0,21*** | ***0,31*** | ***0,23*** | ***0,38*** | ***0,19*** | ***0,26*** | ***0,14*** | **-** |

**Table 6.** Evaluation scores given to the i-th forecaster, split for criterion, averaged across all raters: the element (j,k) of the table is computed as $s_i^{(Ck)} = \frac{1}{16}\sum_{j=1}^{16} \quad s_{ij}^{(Ck)} \in R^{16\times 9}$ .

| **Forecaster** | **C1** | **C2** | **C3** | **C4** | **C5** | **C6** | **C7** | **C8** | **C9** | ***Avg F*** |
|---|---|---|---|---|---|---|---|---|---|---|
| **gpt-4o** | 4,43 | 4,25 | 4,50 | 3,87 | 4,00 | 3,81 | 4,37 | 4,37 | 4,31 | *4,21* |
| **claude-3-5-sonnet** | 4,31 | 4,25 | 4,62 | 4,37 | 4,12 | 3,81 | 4,62 | 4,50 | 4,19 | *4,31* |
| **gemini-1-5-pro-api** | 4,12 | 3,87 | 4,37 | 3,87 | 3,50 | 3,50 | 4,37 | 4,25 | 3,87 | *3,97* |
| **Yi-Large-preview** | 4,68 | 3,87 | 4,68 | 4,43 | 3,75 | 4,18 | 4,56 | 4,31 | 4,31 | *4,31* |
| **Gemma-2-27B** | 4,12 | 4,25 | 4,37 | 4,06 | 3,62 | 4,00 | 3,94 | 4,00 | 3,94 | *4,03* |
| **GLM-4-0520** | 4,37 | 4,00 | 4,43 | 4,25 | 3,93 | 4,06 | 4,44 | 3,94 | 4,00 | *4,16* |
| **Llama-3-70b-Instruct** | 4,37 | 4,25 | 4,50 | 4,31 | 4,12 | 4,31 | 4,69 | 4,25 | 4,12 | *4,33* |
| **Reka-Core-2** | 4,25 | 3,87 | 4,56 | 4,00 | 3,75 | 3,87 | 4,06 | 3,94 | 3,81 | *4,01* |
| **Command-R+** | 4,31 | 4,37 | 4,25 | 4,18 | 4,06 | 4,25 | 4,12 | 4,06 | 3,87 | *4,17* |
| **Qwen2-72B-Instruct** | 4,62 | 4,25 | 4,75 | 4,25 | 4,25 | 4,31 | 4,50 | 4,12 | 4,31 | *4,37* |
| **DeepSeek-Coder-V2** | 4,37 | 3,93 | 4,25 | 4,06 | 4,00 | 4,00 | 4,44 | 4,06 | 3,87 | *4,11* |
| **Mistral-Large-2402** | 4,37 | 4,18 | 4,62 | 4,12 | 3,93 | 4,06 | 4,44 | 4,00 | 3,94 | *4,19* |
| **Mixtral-8x22b-Instruct** | 4,31 | 4,37 | 4,50 | 4,12 | 4,25 | 4,44 | 4,06 | 3,81 | 3,87 | *4,19* |
| **Phi-3-Medium-4k-Instruct** | 4,25 | 3,81 | 4,43 | 4,18 | 4,12 | 3,94 | 4,44 | 4,19 | 4,06 | *4,16* |
| **DBRX-Instruct-Preview** | 4,37 | 4,25 | 4,56 | 4,31 | 3,93 | 4,44 | 4,44 | 4,00 | 4,06 | *4,26* |
| **pplx-70b-online** | 4,87 | 4,00 | 4,93 | 4,31 | 4,37 | 4,44 | 4,75 | 4,44 | 4,44 | *4,51* |
| ***Average C score*** | *4,38* | *4,11* | *4,52* | *4,18* | *3,98* | *4,09* | *4,39* | *4,13* | *4,06* | *4,21* |
| ***Std deviation C score*** | *0.190* | *0.189* | *0.174* | *0.62* | *0.229* | *0.261* | *0.226* | *0.192* | *0.191* | *0,14* |

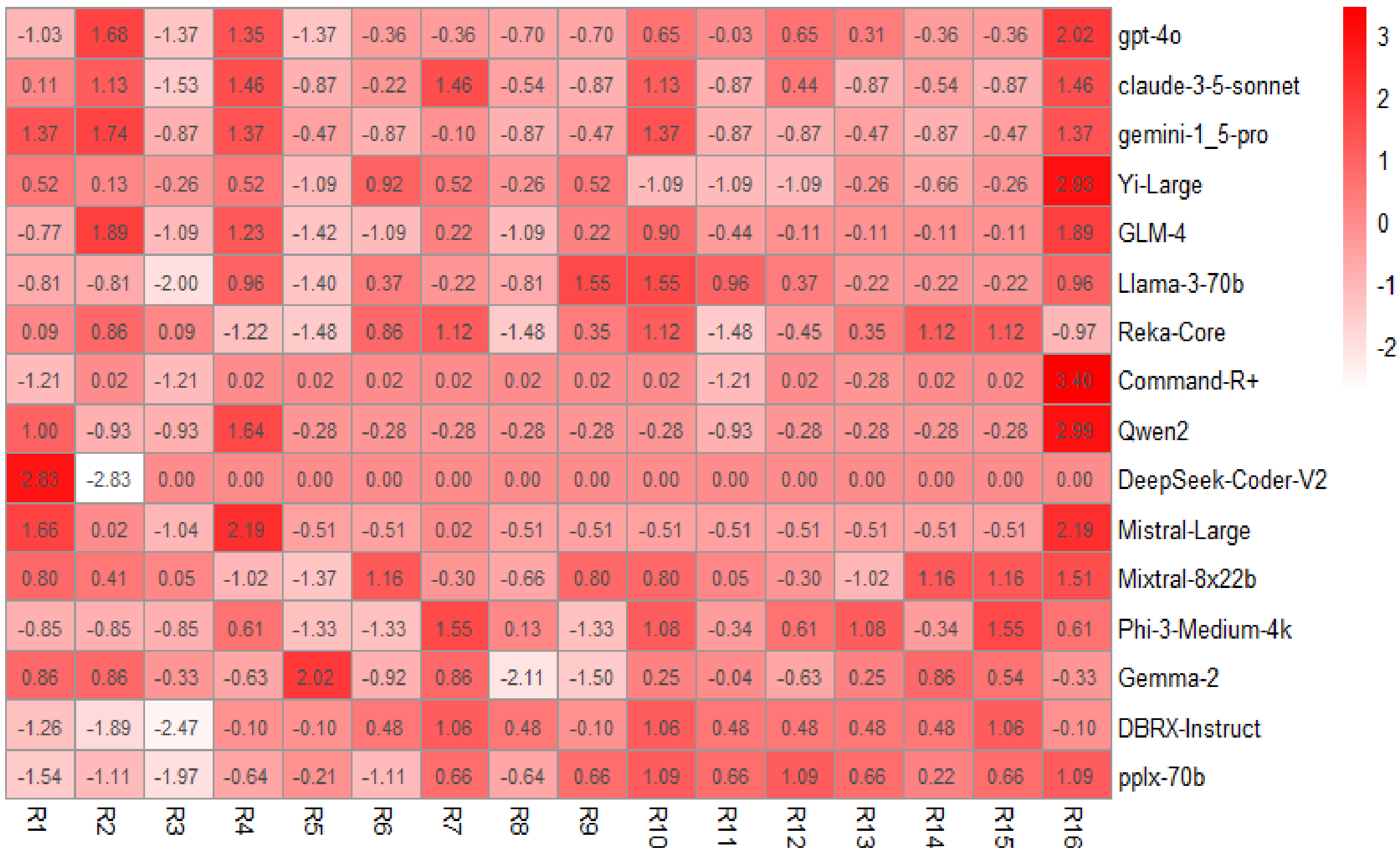

**Figure 1.** Heatmap of the studentized residuals of the scores for the forecasters (on the vertical axis) vs the raters (on the horizontal) - ordering remains the same (e.g. X1 stands for Gpt-4o and X16 for pplx-70b).

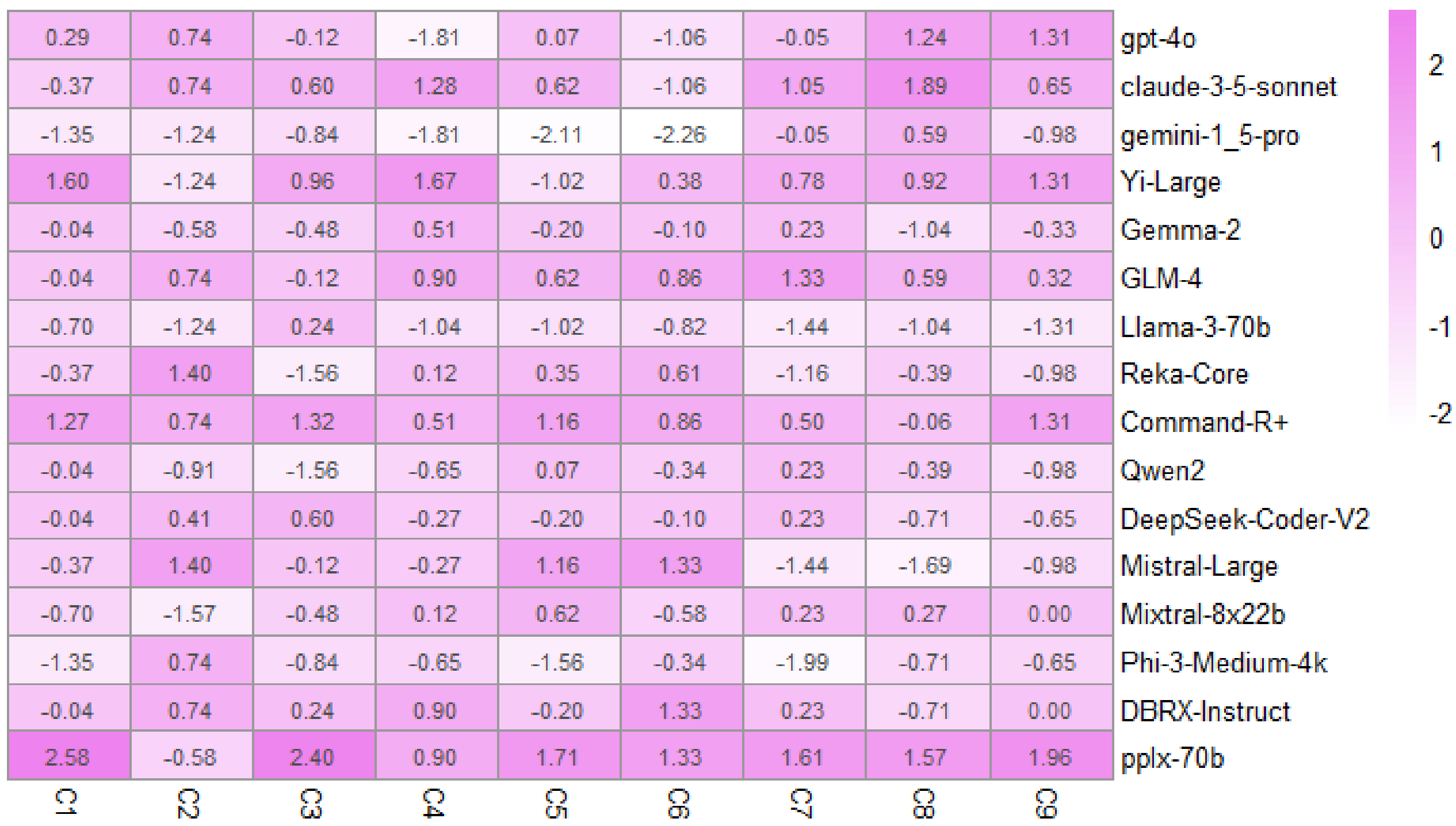

**Figure 2.** Heat map of the studentized residuals of the scores for the forecasters (on the vertical axis) vs the criterion (on the horizontal axis).

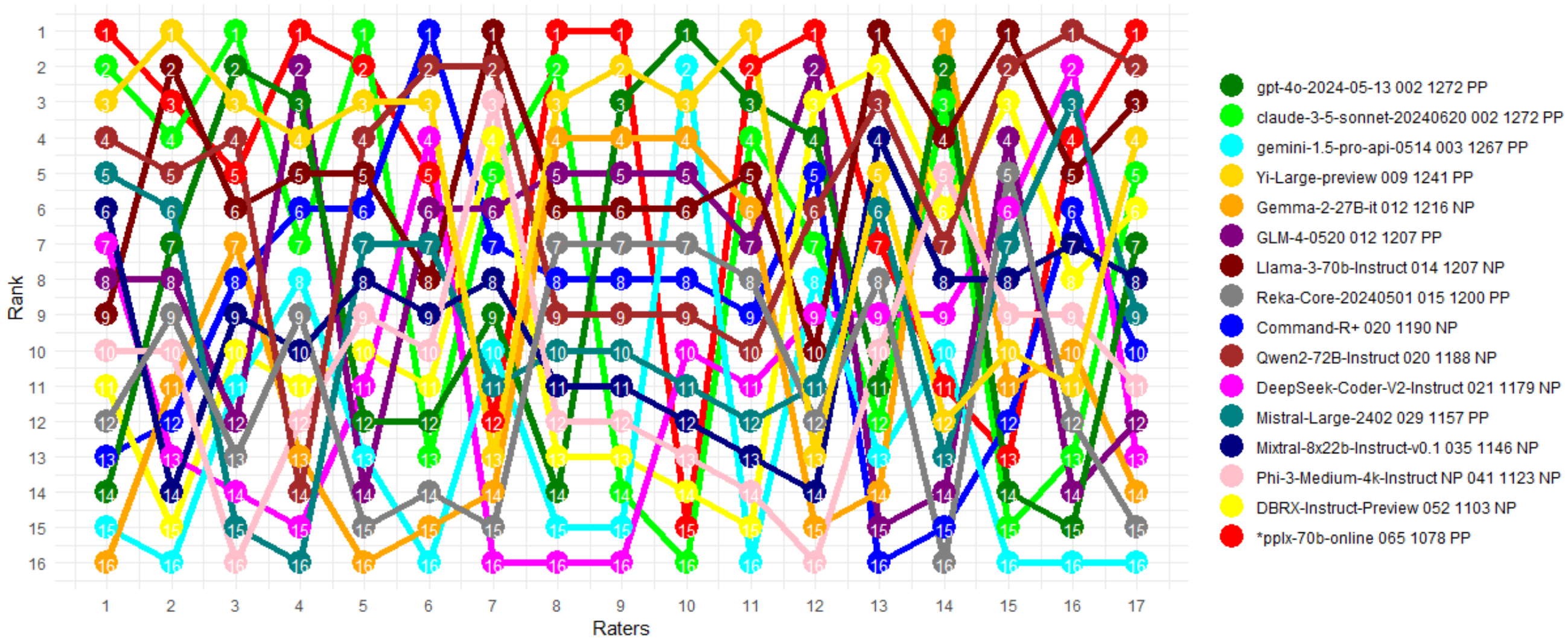

**Figure 3.** Bump chart of rankings, displaying rankings produced by each individual rater, along with the final ranking (rightmost), calculated using uniform weighting of the raters' evaluations. The ranking indices are reported in the last column of Table 5.

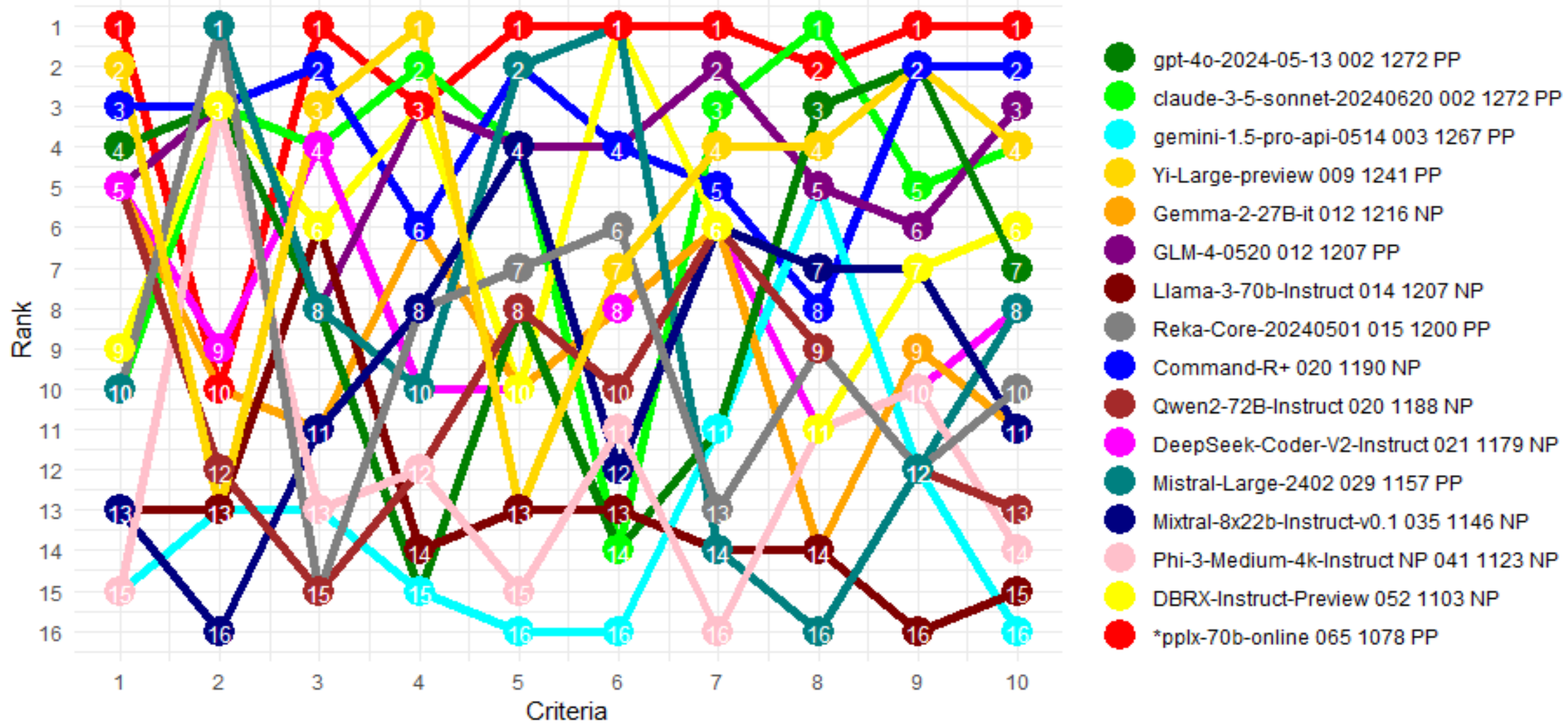

**Figure 4.** Bump chart displaying forecasters rankings produced by the criteria (from 1 to 9) and the uniform score ranking (10)

## 6. Analysis of LLM peer review: agreement and consistency

### *6.1 Inter-rater consistency analysis*

We assessed the consistency of peer review evaluations across different raters using the Intraclass Correlation Coefficient (ICC), a statistical measure that evaluates the level of consistency and agreement among raters. The ICC is a ratio of covariance to total variance, accounting for various sources of variance in the score matrix *S*, including the selection of forecasters and raters. Given the systematic variation in scores between raters, we employed a two-way random model to accurately represent the data. The ICC values were calculated following the definitions provided by McGraw (1996):

- ICC(C,n) estimates the squared correlation of average scores and universe scores, representing the degree of consistency for scores that are averages of n independent ratings on randomly selected forecasts.

- ICC(A,n) estimates the squared correlation of average scores and universe scores, including variance between raters, representing the degree of absolute agreement for scores that are averages based on n independent raters on randomly selected forecasts.

In our analysis, consistency is contrasted with absolute agreement when measuring correlation. In consistency, the total score variance (i.e., differences in the overall scores given by raters) is used as the denominator (McGraw, 1996). If two raters' scores can be aligned by applying an additive transformation (for instance, subtracting each rater's mean score from their individual ratings), they achieve consistency—meaning they rank items the same way—without necessarily agreeing on the exact scores. This distinction explains why agreement-based measures, which require identical scores, tend to be lower than consistency-based measures, which only require the same rankings.

The ICCs for the peer review scores are presented in Table 7. The single intraclass indices, which measure correlations between individual raters' scores, are 4 to 10 times lower than the average intraclass indices, calculated by correlating the average scores from each rater. These average indices are especially relevant since each forecaster is ranked based on a weighted average of the raters' scores. In the random model, this is seen as a combination of independent assessments on randomly selected items. The notable difference between ICC(C,1) and ICC(A,1) is expected due to the large mean differences between groups. The ICC(C,16) value of 0.79 indicates a high level of consistency in the LLM evaluations, reflecting strong agreement within the panel.

**Table 7.** Intraclass Correlation Coefficient (ICC) computed for various combinations of unit and type, based on a two-way statistical model.

| ICC Model/Type | ICC | 95% confid. interval | F-Test |
|---|---|---|---|
| Single Score Intraclass Correlation Type: consistency | ICC(C,1) = 0.199 | [0.093; 0.41] | F(15,22)= 4.98 p = 2e-08 |
| Average Score Intraclass Correlation Type: consistency | ICC(C,16) = 0.799 | [0.620; 0.917] | F(15,22) = 4.98 p = 2e-08 |
| Single Score Intraclass Correlation Type: agreement | ICC(A,1) = 0.0417 | [0.013; 0.116] | F(15,33) = 4.98 p = 5.88e-05 |
| Average Score Intraclass Correlation Type: agreement | ICC(A,16) = 0.410 | [0.148, 0.687] | F(15,22)= 4.98 p = 0.000369 |

To determine if the ICC(C,16) value is significantly greater than zero, we tested the hypothesis that the correlation is higher than what would be expected for a medium-sized effect. The F-statistics allowed us to reject the null hypothesis, with a p-value well below the 5% significance threshold. This result confirms the validity of the peer review process and indicates that LLMs can reliably assess each other's performance based on the established criteria.

**Table 8.** Intraclass Correlation Coefficient (ICC) computed for various combinations of unit and type, for data in Table 6, based on a two-way statistical model.

| ICC Model/Type | ICC | 95% confidence interval | F-Test |
|---|---|---|---|
| Single Score Intraclass Correlation Type: consistency | ICC(C,1) = 0.377 | [0.204; 0.623] | F(15,120) = 6.44 p = 6.95e-10 |
| Average Score Intraclass Correlation Type: consistency | ICC(C,9) = 0.845 | [0.698; 0.937] | F(15,120) = 6.44 p = 6.95e-10 |
| Single Score Intraclass Correlation Type: agreement | ICC(A,1) = 0.221 | [0.088; 0.453] | F(15,32.9) = 6.44 p = 4.39e-06 |
| Average Score Intraclass Correlation Type: agreement | ICC(A,9) = 0.718 | [0.44; 0.884] | F(15,25.3) = 6.44 p = 2.32e-05 |

In Table 8 the total score variance is computed differences in the overall scores given per criterion. The single intraclass indices, which measure correlations between criteria' scores, are greater than in the raters' scores but still too low to extract any conclusion. The ICC(C,9) value of 0.84 indicates a high level of consistency in the criteria evaluations, meaning that all criteria contribute constructively to the total variance and the final ranking of the LLMs.

*6.2 Alternative ranking methods*

In the previous section, we analyzed the consistency of peer review scores using standard ranking methods based on uniform weighting, if all raters contribute equally to the final rankings. However, given that raters can vary in their expertise and accuracy across different tasks, it is worth exploring whether alternative ranking methods—especially those incorporating external benchmarks—could yield different results. In this section, we investigate the impact of applying weights to the raters' scores based on their performance in external evaluations, such as the LLM becnhmarks. By adjusting the peer review scores using these benchmarks, we aim to assess whether the relative rankings of the forecasters change and whether the models' performance in forecasting AGI events can be linked to their broader capabilities, as measured by external evaluations. For this purpose, we selected three diverse benchmarks: LMSYS Chatbot Arena, MixEval, and AlpacaEval (updated as of July 17, 2024) whose benchmark values are in Table 9. This selection provides a diversified benchmark portfolio, assessing a wide range of competencies and methodologies. Using the LLMs' scores in these benchmarks (as shown in Table 8), we developed a weighting system to adjust their scores. Specifically, we consider the benchmark scores of the ith rater, denoted as $b_i^{Ar}$ for Arena, and apply the L1 normalization to obtain the weights for equation (1):

$$w_i^{Ar} = \frac{b_i^{Ar}}{\sum_{l=1}^{16} b_l^{Ar}} \quad (2)$$

Table 10 presents the evaluation scores of the forecasters, calculated by weighting the raters using uniform weights and the three selected benchmarks. The resulting scores show significant differences. Figure 5 illustrates the rankings derived from the scores in Table 9.

It could be expected that using different evaluation panels based on diverse benchmarks should result in significantly different rankings of the LLMs. However, contrary to this prediction, the rankings are surprisingly similar. This is evident in the top five LLMs, which remain the same across all rankings (as per Fig. 3, columns from 1 to 4). Further, Table 14 shows high similarity between the four rankings in discourse, as measured by their normalized Kendall distance, whose maximum is 0.133.

The main conclusion from these findings is that the choice of benchmark used to evaluate the LLMs does not significantly impact the final rankings. We suggest two possible explanations for this result. Either the used benchmarks fully capture the skills needed for AGI forecasting, meaning that they are all measuring the right competencies, or the benchmarks do not assess the necessary abilities at all, implying that none are truly relevant for this specific task. This is an important observation because it suggests that either standard benchmarks are perfectly aligned with the skills required for AGI forecasting, or that new evaluation methods, specifically designed for this complex and speculative task, may be needed.

**Table 9.** The benchmark values of LLMs as per Chatbot Arena, MixEval, and AlpacaEval (na = value not available; * = the closest available value in the same LLM family).

| LLMs | Arena | Mix Eval | Alpaca Eval |
|---|---|---|---|
| *gpt-4o* | 1282 | 87,9 | 57,5 |
| *claude-sonnet* | 1272 | 89,9 | 40,5 |
| *gemini-1.5* | 1267 | 84,2 | 24,4 |
| *Yi-Large* | 1241 | 84,4 | 51,9 |
| *GLM-4* | 1216 | 69,6 | 10,4 |
| *Llama-3-70b* | 1207 | na | na |
| *Reka-Core* | 1207 | 84,0 | 34,4 |
| *Command-R+* | 1200 | 83,4 | na |
| *Qwen2-72B* | 1190 | 81,5 | 10,9 |
| *DeepSeek-Coder* | 1188 | 86,1 | 36,6 |
| *Mistral* | 1179 | 83,7 | na |
| *Mixtral* | 1157 | 84,3 | 32,7 |
| *Phi-3-Medium* | 1146 | 76,4 | 30,9 |
| *Gemma-2* | 1123 | na | 7,8 |
| *DBRX* | 1103 | na | 25,4 |
| *pplx-70b* | 1078 | na | na |

**Table 10**. Evaluation scores calculated by weighting raters uniformly or according to their benchmark values.

| LLMs | Uniform | Arena | Mix Eval | Alpaca Eval |
|---|---|---|---|---|
| *gpt-4o* | 4,22 | 4,20 | 4,64 | 4,62 |
| *claude-sonnet* | 4,31 | 4,31 | 4,78 | 4,82 |
| *gemini-1.5* | 3,97 | 3,96 | 4,37 | 4,38 |
| *Yi-Large* | 4,31 | 4,31 | 4,76 | 4,75 |
| *GLM-4* | 4,04 | 4,01 | 4,29 | 4,31 |
| *Llama-3-70b* | 4,16 | 4,15 | 4,56 | 4,64 |
| *Reka-Core* | 4,33 | 4,31 | 4,7 | 4,78 |
| *Command-R+* | 4,01 | 4,00 | 4,39 | 4,42 |
| *Qwen2-72B* | 4,17 | 4,15 | 4,54 | 4,55 |
| *DeepSeek-Coder* | 4,38 | 4,36 | 4,75 | 4,81 |
| *Mistral* | 4,11 | 4,09 | 4,39 | 4,44 |
| *Mixtral* | 4,19 | 4,17 | 4,54 | 4,59 |
| *Phi-3-Medium* | 4,20 | 4,18 | 4,53 | 4,59 |
| *Gemma-2* | 4,16 | 4,14 | 4,47 | 4,52 |
| *DBRX* | 4,26 | 4,24 | 4,62 | 4,68 |
| *pplx-70b* | 4,51 | 4,5 | 5 | 5 |

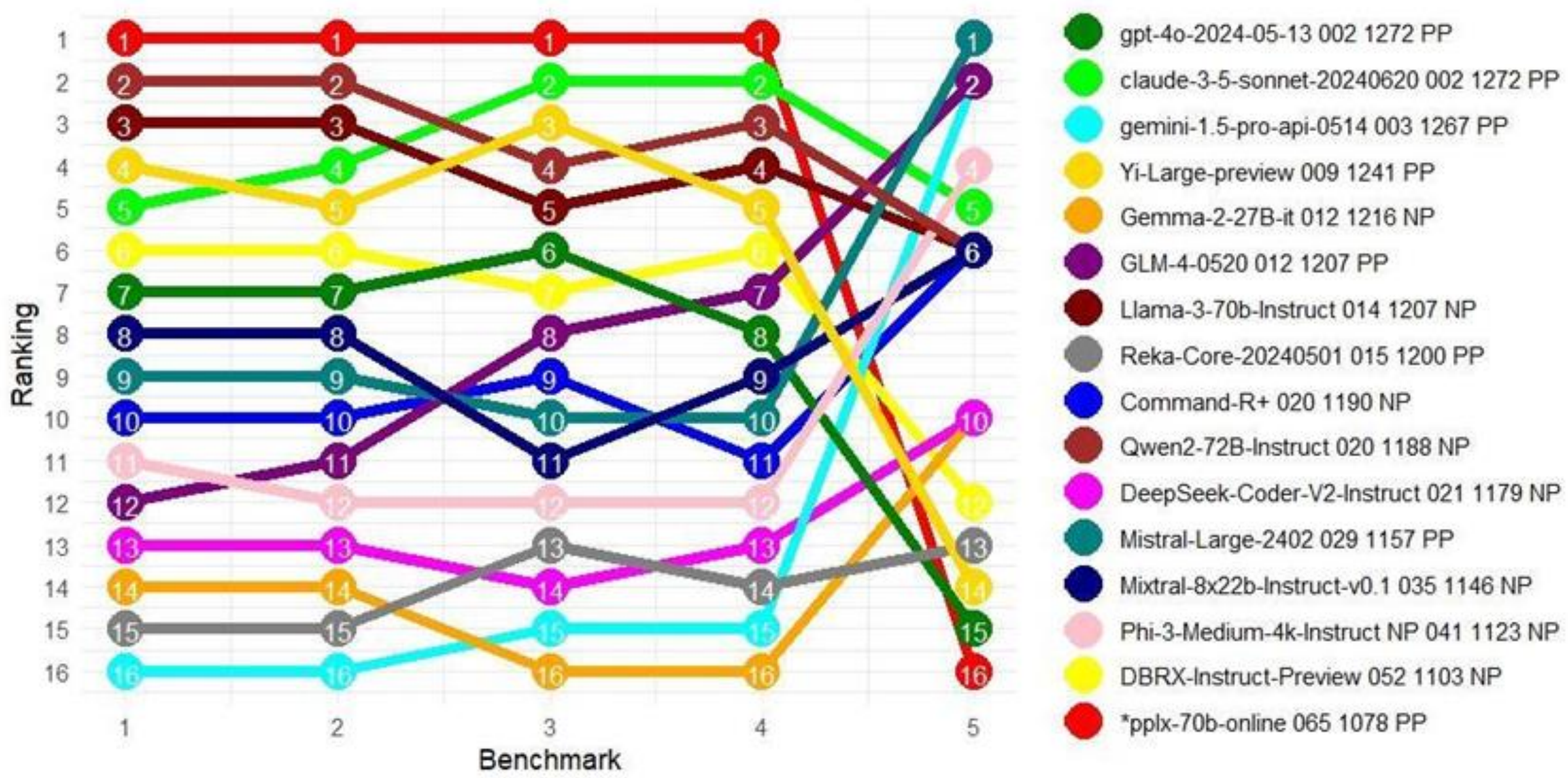

**Figure 5.** Bump chart of rankings, respectively induced by the benchmarks: (1) Uniform, (2) Arena, (3) MixEval, (4) AlpacaEval, (5) Expert ranking.

### *6.3 Analysis of LLM self-evaluation*

After exploring the impact of different ranking methods in Section 6.2, we now investigate how LLMs assess their own performance (i.e., Self-Evaluation) compared to how they are evaluated by others (i.e., Hetero-Evaluation). By examining the accuracy of self-evaluations, we aim to understand whether certain models tend to overestimate or underestimate their performance. In Table 11, we compare the self-evaluations of each model with the average evaluations they received from others. DeepSeek Coder V2 Instruct and Mistral Large 2402R showed a better balance, with their self-assigned scores closely aligning with the scores given by other LLMs. In contrast, DBRX-Instruct-Preview and Mixtral-8x22b-Instruct-v0.1 displayed significant self-preference, assigning themselves scores that were 17% higher than those from others. Gemini-1.5-pro-api-0514, the most critical in its evaluations of forecasts, demonstrated marked self-underestimation, giving itself a score 40% lower than the average it received from others.Further we define a LLM's Self-Evaluation Index (SEI) as the ratio between its self-assessment score (SES) and the average score it received from other LLMs (hetero-evaluation score - HES):

$$SEI_i = \frac{SES_i}{HES_i} = \frac{s_{ii}}{\sum_{j=1\ j\neq i}^{16} \quad s_{ij}/15} \qquad (3)$$

The SEI indices are presented in Table 11. Values above 1.0 indicate self-overestimation and values below 1.0 indicate self-underestimation (relative to peer evaluations). SEI values range from 0.599 (gemini-1.5-pro-api-0514) to 1.184 (DBRX-Instruct-Preview), with DeepSeek-Coder-V2-Instruct achieving a perfect balance at SEI of 1.0.

**Table 11**. LLMs average self-evaluation score (SES), average Hetero-evaluation Score (HES), Self-Evaluation Index (SEI), and the uniform weighted score for comparison.

| LLM | Uniform | SES | HES | SEI |
|---|---|---|---|---|
| gpt-4o-2024-05-13 | 4,21 | 3,89 | 4,24 | 0,92 |
| claude-3-5-sonnet-20240620 | 4,31 | 4,56 | 4,30 | 1,06 |
| gemini-1.5-pro-api-0514 | 3,97 | 2,44 | 4,07 | 0,60 |
| Yi-Large-preview | 4,31 | 3,89 | 4,34 | 0,90 |
| Gemma-2-27B-it | 4,03 | 2,89 | 4,11 | 0,70 |
| GLM-4-0520 | 4,16 | 4,22 | 4,16 | 1,02 |
| Llama-3-70b-Instruct | 4,33 | 5,00 | 4,28 | 1,17 |
| Reka-Core-20240501 | 4,01 | 4,56 | 3,98 | 1,15 |
| Command-R+ | 4,17 | 4,11 | 4,17 | 0,99 |
| Qwen2-72B-Instruct | 4,37 | 4,78 | 4,35 | 1,10 |
| DeepSeek-Coder-V2-Instruct | 4,11 | 4,11 | 4,11 | 1,00 |
| Mistral-Large-2402 | 4,19 | 4,22 | 4,18 | 1,01 |
| Mixtral-8x22b-Instruct-v0.1 | 4,19 | 4,89 | 4,15 | 1,18 |
| Phi-3-Medium-4k-Instruct | 4,16 | 4,56 | 4,13 | 1,10 |
| DBRX-Instruct-Preview | 4,26 | 5,00 | 4,22 | 1,19 |
| pplx-70b-online | 4,51 | 5,0 | 4,48 | 1,12 |

Table 11 shows that uniformly weighted scores and the HES are quite identical, as expected. For a more comprehensive analysis, we now carry out a Pearson correlation between SES, HES, SEI, and the Arena scores. As indicated by Table 12 and Fig. 4, a significant negative correlation exists between SES score and Arena value, as well as between SEI score and Arena value (see p-value). This suggests that the higher LLMs scored on Arena, the less they tended to estimate their output. On the contrary the correlation is negligible between Arena on one side and either Uniform score, HES or Expert score on the other side. We also calculated the cosine similarity between the normalized 16-dimensional vectors of the SEI and the Arena values, obtaining a value of 0.99. This result agrees with the results of the correlation analysis shown in Table 12.

**Table 12.** Correlation analysis of the Arena value of the LLMs vs Uniform score, b) Self-Evaluation Score (SES), c) Hetero-Evaluation Score (HES), d) Self-Evaluation Index (SEI), e) Expert score - the latter is introduced in Sect. 7.1.

| | **Intercept** | **Slope** | **Residual standard error** | **Degrees of freedom** | **Adjusted R-squared** | **F-statistic** | **Pearson coefficient** | **p-value** |
|---|---|---|---|---|---|---|---|---|
| **uniform score** | 5.15 | -0.0008 | 0.13 | 14 | 0.0492 | 1.777 | -0.335 | 0.2038 |
| **SES** | 12.87 | -0.0072 | 0.61 | 14 | 0.2950 | 7.277 | -0.585 | **0.0173** |
| **HES** | 4.65 | -0.0003 | 0.12 | 14 | 0.0321 | 0.465 | -0.179 | 0.5063 |
| **SEI** | 2.96 | -0.0016 | 0.14 | 14 | 0.2921 | 7.190 | -0.582 | **0.0179** |
| **Expert score** | 3.63 | 0.0003 | 1.37 | 14 | -0.0711 | 0.003 | 0.016 | 0.9505 |

**Figure 4.** Linear Interpolation of a) SES and b) SEI vs the Arena score.

a)

b)

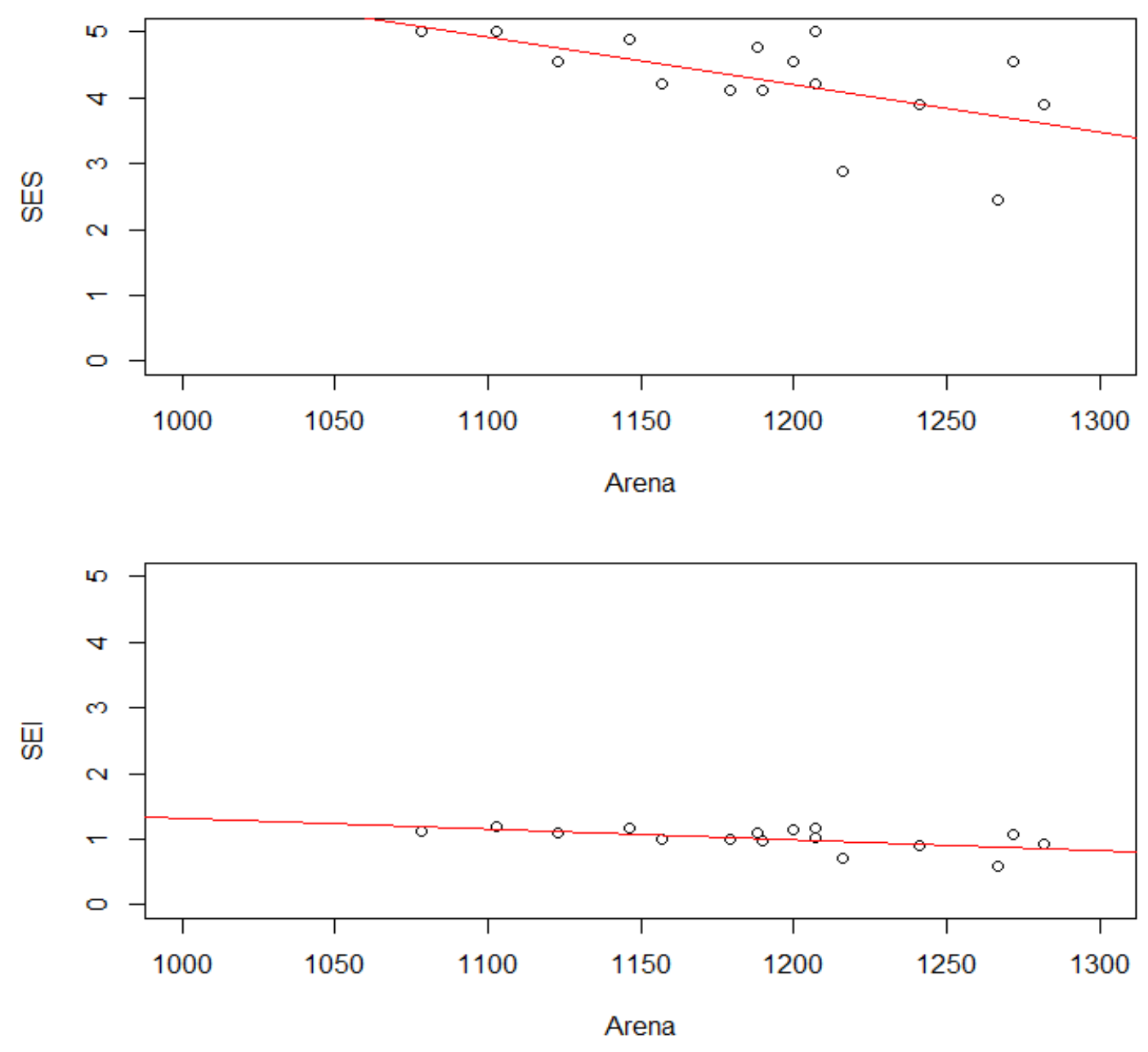

## 7. Comparing LLMs to Human Experts on AGI

### *7.1 Comparing LLMs forecast to the human expert likelihood estimation*

We now shift focus to comparing AGI likelihood estimates from LLMs with those of human AI experts, as reported in Grace et al. (2024). The aggregate expert estimate of AGI likelihood by 2027 is 10%. After an adjustment caused by our term equal to 2030, we consider this as a reference value to assess how closely LLM predictions align with human judgment. Additionally, we explore whether benchmark weighting can enhance this alignment, offering insights into the relationship between LLM performance and the reliability of AGI forecasting. It's important to clarify that here, "reliability" does not refer to absolute accuracy, as AGI prediction is inherently uncertain, even for human experts. In other words, the aim of our assessment method is to understand how LLMs handle uncertain predictions and evaluate them (i.e., the process), rather than how closely they match an objectively correct answer (i.e., the outcome).

First, we computed the simulated scores that human experts would have assigned to the forecasters based solely on their predictions of AGI likelihood (the scoring formula is reported in appendix). The computation starts from a similarity measure and reduces it to the standard 1-5 Likert scale we used in the previous discussion. Table 13 (in the fourth column) reports the simulated scores. Notably, Mixtral-8x22b-Instruct-v0.1, GLM-4-0520 012 1207 PP, and Gemini-1.5-pro-api-0514 show the closest alignment with the adjusted human estimates. This comparative framework allows us to assess the degree of concordance between LLM-generated forecasts and those of human experts, providing an additional external validation metric for evaluating the LLM performance in the complex task of AGI prediction. It is evident that 13 LLMs out of 16 receive scores over 4, showing a good evaluation from the human expert panel.

**Table 13.** Expert ranking of LLMs based on expert-derived scores. The scores are calculated as a function of the LLM estimates and expert estimates of the likelihood of the AGI event (see appendix).

| **Ranking** | **AGI % likelihood by 2030** | **Grace et al. % likelihood by 2027** | **Expert score** |
|---|---|---|---|
| Mistral-Large-2402 | 12 | 10 | 4,98 |
| GLM-4-0520 | 8 | 10 | 4,94 |
| Gemini-1.5-pro-api-0514 | 12,5 | 10 | 4,94 |
| Phi-3-Medium-4k-Instruct | 6,3 | 10 | 4,75 |
| claude-3-5-sonnet-20240620 | 5,8 | 10 | 4,70 |
| Llama-3-70b-Instruct | 15 | 10 | 4,66 |
| Command-R+ | 15 | 10 | 4,66 |
| Qwen2-72B-Instruct | 15 | 10 | 4,66 |
| Mixtral-8x22b-Instruct-v0.1 | 15 | 10 | 4,66 |
| Gemma-2-27B-it | 5 | 10 | 4,61 |
| DeepSeek-Coder-V2-Instruct | 5 | 10 | 4,61 |
| DBRX-Instruct-Preview | 3,5 | 10 | 4,44 |
| Reka-Core-20240501 | 3 | 10 | 4,38 |
| Yi-Large-preview | 38 | 10 | 2,08 |
| gpt-4o-2024-05-13 | 45 | 10 | 1,29 |
| pplx-70b-online | 47,6 | 10 | 1,00 |

### *7.2 Evaluation performance*

Figure 5, in the fifth column, shows the expert ranking, that is based on the expert simulated scores (Table 13) and compares it to the other rankings shown in Figure 5 (which were based on benchmark weights). Crucially, the expert ranking shows a dramatic reshuffling of the LLMs' positions. In fact, 4 out of 5 LLMs (80%) in the top group have now changed ranking. Only one LLM, Claude-3-5-sonnet-20240620, remains in the top 5 in both the previous rankings and this new human-aligned ranking. This indicates a significant difference between how LLMs align with human experts in evaluation of AGI. A more quantitative perspective is provided by the Kendall normalized distances between the five rankings shown in Figure 5. The Arena and Uniform benchmarks have the smallest distance between the raters (1,6%), indicating that using the Arena benchmark yields similar results to applying equal weighting to the raters. However, both rankings show a substantial distance from the Expert ranking (approximately 0.7). This suggests that the Arena benchmark is unlikely to help the panel align its evaluation with expert judgment. A similar pattern emerges with the MixEval and Alpaca benchmarks. In conclusion, within our evaluation framework, performance on standard AI benchmarks does not correlate strongly with AGI predictions that match expert opinions. This underscores the uniqueness of the AGI prediction task and suggests that different skills or capabilities might be required for this specific type of forecasting compared to those measured by standard AI benchmarks.

**Table 14.** Kendall normalized distances between the rankings generated by a) Uniform, b) Arena, c) MixEval, d) AlpacaEval, e) Expert, f) AGI 16 Bench.

| | Uniform | Arena | MixEval | Alpaca | Expert | AGI 16 Bench |
|---|---|---|---|---|---|---|
| **Uniform** | 0 | 0.0167 | 0.1333 | 0.0833 | 0.575 | 0.3167 |
| **Arena** | 0.0167 | 0 | 0.1167 | 0.0667 | 0.5583 | 0.3167 |
| **MixEval** | 0.1333 | 0.1167 | 0 | 0.0667 | 0.5667 | 0.3833 |
| **Alpaca** | 0.0833 | 0.0667 | 0.0667 | 0 | 0.525 | 0.3333 |
| **Expert** | 0.575 | 0.5583 | 0.5667 | 0.525 | 0 | 0,3583 |
| **AGI 16 Bench** | 0.3167 | 0.3167 | 0.3833 | 0.3333 | 0,3583 | 0 |

*7.3 Confidence weight of raters*

According to Ning et al. (2024) we considered evaluating $w_j$ in equation (1) as a confidence weight for the j-th rater. As the peer-review process works in an unsupervised way, we adapted confidence weights to obtain a ranking closer to the expert ranking. It is a constrained optimization, where score matrix S is constant, while the confidence vector is varied to adjust the final scores and the ranking to align with the expert scores and ranking respectively (Table 13). Here, we are not making strong assumptions - as in Ning et al. (2024) - that high-level LLM can evaluate forecasters more accurately than low-level ones, while at the same time this pretended higher-level LLMs achieve higher scores as forecasters.

Let us assume as reference the expert ranking $R^{(exp)}$, (Table 13), which aligns with human preferences and is determined by the expert scores $s_i^{(exp)}$ (Table 13, 4th column). We therefore look for the ranking $\hat{R}$, estimated through the optimization process, which minimizes an appropriate loss function $L\left(R^{(exp)}, \hat{R}\right)$ to get as close as possible to the human ranking $R^{(exp)}$. To this end we adopt the normalized Kendall distance $\tau_K$ computed between the two rankings, and state:

$$argmin\ L\left(R^{(exp)}, \hat{R}\right)$$
$$\text{s.t.} \quad \sum_{j=1}^{16} \hat{w}_j = 1 \ and\ \hat{w}_j > 0$$
$$L\left(R^{(exp)}, \hat{R}\right) = \alpha \sum_{i=1}^{16} \left(s_i^{(exp)} - \hat{s}_i\right)^2 + \beta \tau_K\left(R^{(exp)}, \hat{R}\right) \qquad (4)$$

where $\hat{s}_i = \sum_{j=1}^{16} \hat{w}_j S_{ij}$ and the hyperparameters α, β take only positive values. As a starting point for the optimization, we took the arena ranking with $\tau_K\left(R^{(exp)}, R^{(arena)}\right) = 0.5583$. The results are shown in Table 15, organized per optimization procedure and hyperparameters - e.g alabama(1,76).

After optimization some LLMs obtain a zero-confidence weight. This means they are excluded from the evaluation panel and do not contribute to the counting function any longer. The various panels that result after each combination of optimization procedure and hyperparameters are different in size and composition. The best panels reduce to 2-7 raters, i.e. those with confidence greater than 10% of the maximum value. The raters that are more represented in the panels are: Gemini-1.5 Pro, Llama-3-70b-Instruct, DBRX-Instruct-Preview, pplx-70b-online. Among the excluded raters, i.e. those with the lowest confidence, the most penalized are Gpt-4o, Yi Large, Gemma2, Reka. Unfortunately, all results are sub-optimal, as neither the Kendall distance nor residuals cancel. An acceptable trade-off satisfies one or both the conditions (that are straightforward as we start from the arena ranking and apply regularization):

$$\tau_K\left(R^{(exp)}, \hat{R}\right) < \tau_K\left(R^{(exp)}, R^{(arena)}\right)$$
$$\tau_K\left(R^{(exp)}, \hat{R}\right) \leq \tau_K\left(R^{(arena)}, \hat{R}\right)$$

(5)

Thus, an acceptable result is given by L-BFGS-B (1,73), with a panel consisting of GLM-4-0520, Llama-3-70b-Instruct, Phi-3-Medium-4k-Instruct, DBRX-Instruct-Preview, Pplx-70b-online. This suboptimal panel produces a ranking with three LLMs in the same position as in the expert ranking (Yi-Large-preview as 4-th, Qwen2-72B-Instruct as 10-th, Phi-3-Medium-4k-Instruct as 14-th), being at a low normalized Kendall distance from the expert ranking (0.358) and determining low quadratic residuals (27.81) as shown in Figure 6. Table 14 shows the fulfillment of one of eqnn (5).

Considering that Gpt-4o and pplx-70b-online are strongly penalized by the expert scores, as shown in Table 13, and that they are far last in the ranking (respectively 15th and 16th), we can consider them as outliers and exclude them as forecasters and raters from the optimization procedure. Results are shown in Table 16. According to the same criteria, another suboptimal solution comes from DEoptim(1,17), with a panel composed of Claude 3.5 Sonnet, Llama-3-70b-Instruct, Reka-Core, Qwen2 72B, Mistral-Large-2402, Phi-3-Medium-4k-Instruct, DBRX-Instruct- Preview. LLama 3 results 6-th in the forecasters ranking, in the same position it holds in the expert ranking. DEoptim(1,17) realizes a lower normalized Kendall distance from the expert ranking (0,308) and very low quadratic residuals (6,1) (Figure 7).

**Table 15.** Results of the optimization process for panel size equal to 16. Legenda: $\hat{R}$ is the optimization Solution; Kendall Distance between $\hat{R}$ and $R^{(exp)}$ is KD_E and NKD_E when we adopt the normalized Kendall Distance; number of coincidences between $R^{(exp)}$ and $\hat{R}$ is C_E; Kendall Distance between $\hat{R}$ and $R^{(arena)}$ is KD_A and NKD_A; number of coincidences between $R^{(arena)}$ and $\hat{R}$ $is$ C_A. Panel size is the number of LLMs with non-zero confidence weight; Panel reports the index of LLMs with non-zero confidence weight.

| **Library / Function** | **Alpha** | **Beta** | **KD_E** | **NKD_E** | **C_E** | **KD_A** | **NKD_A** | **C_A** | **Panel size after opt.** | **Panel** | **Quadratic residues** |
|---|---|---|---|---|---|---|---|---|---|---|---|
| alabama | 1 | 76 | 41 | 0,342 | 1 | 40 | 0,333 | 5 | 3 | 3,7,14 | 25,6114 |
| alabama | 1 | 1 | 44 | 0,367 | 0 | 36 | 0,3 | 1 | 2 | 3,7 | 25,2791 |
| COBYLA | 1 | 1 | 49 | 0,408 | 3 | 31 | 0,258 | 0 | 7 | 3,5,6,7, 14,15, 16 | 26,4304 |
| COBYLA | 1 | 60 | 61 | 0,508 | 0 | 6 | 0,05 | 5 | 4 | 2,7,12, 14 | 29,5467 |
| constr Optim | 1 | 1 | 41 | 0,342 | 3 | 40 | 0,333 | 2 | 3 | 7,15,16 | 23,8163 |
| constr Optim | 1 | 70 | 42 | 0,35 | 2 | 36 | 0,3 | 2 | 3 | 7,15,16 | 24,9178 |
| Deoptim | 1 | 69 | 42 | 0,35 | 1 | 38 | 0,317 | 1 | 3 | 3,6,7 | 25,6373 |
| Deoptim | 1 | 0 | 44 | 0,367 | 0 | 36 | 0,3 | 1 | 2 | 3,7 | 25,2845 |
| Deoptim | 1 | 1 | 44 | 0,367 | 0 | 36 | 0,3 | 1 | 2 | 3,7 | 25,2841 |
| Deoptim | 0 | 1 | 48 | 0,4 | 3 | 23 | 0,192 | 4 | 3 | 1,7,16 | 0 |
| genSA | 1 | 80 | 41 | 0,342 | 2 | 38 | 0,317 | 1 | 3 | 7,15,16 | 27,5877 |
| genSA | 1 | 1 | 44 | 0,367 | 1 | 39 | 0,325 | 1 | 2 | 6,15 | 26,8383 |
| **L-BFGS-B** | **1** | **73** | **43** | **0,358** | **3** | **38** | **0,317** | **1** | **5** | **6,7,14, 15,16** | **27,8055** |
| L-BFGS-B | 1 | 1 | 44 | 0,367 | 0 | 36 | 0,3 | 1 | 2 | 3,7 | 25,2791 |
| MLSL | 1 | 60 | 42 | 0,35 | 1 | 42 | 0,35 | 1 | 1 | 7 | 26,6122 |
| MLSL | 1 | 1 | 44 | 0,367 | 2 | 38 | 0,317 | 1 | 2 | 7,16 | 26,6508 |
| Nelder-Mead | 1 | 1 | 60 | 0,5 | 1 | 15 | 0,125 | 3 | 5 | 5,7,14,15,16 | 28,0067 |
| Nelder-Mead | 1 | 56 | 63 | 0,525 | 0 | 4 | 0,033 | 10 | 3 | 6,15,16 | 29,3595 |
| PRAXIS | 1 | 60 | 43 | 0,358 | 0 | 37 | 0,308 | 2 | 2 | 5,7 | 25,4761 |
| PRAXIS | 1 | 1 | 45 | 0,375 | 0 | 37 | 0,308 | 2 | 3 | 3,7,15 | 25,4006 |
| PSO | 1 | 60 | 42 | 0,35 | 1 | 38 | 0,317 | 1 | 2 | 7,15 | 27,6005 |
| PSO | 1 | 1 | 43 | 0,358 | 1 | 38 | 0,317 | 0 | 1 | 7 | 26,7417 |
| Rsolnp | 1 | 1 | 44 | 0,367 | 0 | 37 | 0,308 | 2 | 2 | 3,7 | 25,2791 |
| Rsolnp | 1 | 69 | 44 | 0,367 | 0 | 37 | 0,308 | 2 | 2 | 3,7 | 25,2791 |

**Table 16.** Results of the optimization process for the reduced panel with 14 LLMs. Legenda: as in Table 15

| Library / Function | Alpha | Beta | KD_E | NKD_E | C_E | KD_A | NKD_A | C_A | Panel size after opt. | Panel | Quadratic residues |
|---|---|---|---|---|---|---|---|---|---|---|---|
| alabama | 1 | 1 | 32 | 0,352 | 1 | 26 | 0,286 | 2 | 4 | 7,10,12,14 | 5,5945 |
| alabama | 1 | 16 | 33 | 0,362 | 1 | 26 | 0,286 | 0 | 4 | 7,10,12,14 | 5,6427 |
| COBYLA | 1 | 1 | 33 | 0,363 | 1 | 25 | 0,275 | 2 | 4 | 7,10,12,14 | 5,5944 |
| COBYLA | 1 | 17,5 | 35 | 0,385 | 0 | 18 | 0,198 | 0 | 4 | 4,7,10,13 | 7,0943 |
| constr Optim | 1 | 1 | 34 | 0,374 | 1 | 19 | 0,209 | 1 | 6 | 7,8,10,12,14,15 | 6,6505 |
| constr Optim | 1 | 18 | 40 | 0,44 | 0 | 11 | 0,121 | 1 | 4 | 7,10,11,14 | 7,5444 |
| **Deoptim** | **1** | **17** | **29** | **0,319** | **0** | **19** | **0,209** | **3** | **7** | **2,7,8,10,12,14,15** | **6,6066** |
| Deoptim | 0 | 1 | 28 | 0,308 | 1 | 24 | 0,264 | 4 | 2 | 7,12 | 0 |
| Deoptim | 1 | 1 | 33 | 0,363 | 1 | 25 | 0,275 | 2 | 4 | 7,10,12,14 | 5,631 |
| Deoptim | 1 | 0 | 34 | 0,374 | 0 | 24 | 0,264 | 2 | 4 | 7,10,12,14 | 5,6118 |
| GenSA | 1 | 1 | 9 | 0,099 | 0 | 11 | 0,121 | 1 | 1 | 10 | 6,7713 |
| GenSA | 1 | 27 | 34 | 0,374 | 2 | 25 | 0,275 | 1 | 4 | 4,5,10,12 | 10,1982 |
| L-BFGS-B | 1 | 1 | 32 | 0,352 | 1 | 26 | 0,286 | 2 | 4 | 7,10,12,14 | 5,5947 |
| L-BFGS-B | 1 | 17 | 33 | 0,363 | 2 | 21 | 0,231 | 1 | 7 | 7,8,10,12,13,14,15 | 6,1835 |
| MLSL | 1 | 1 | 34 | 0,374 | 2 | 24 | 0,264 | 3 | 3 | 7,10,14 | 6,0745 |
| MLSL | 1 | 17 | 34 | 0,374 | 2 | 22 | 0,242 | 2 | 1 | 7 | 6,1525 |
| Nelder-Mead | 1 | 17 | 40 | 0,44 | 0 | 9 | 0,099 | 5 | 2 | 7,8,11 | 7,8856 |
| Nelder-Mead | 1 | 1 | 42 | 0,462 | 0 | 14 | 0,154 | 3 | 5 | 7,9,10,12,15 | 7,1607 |
| PRAXIS | 1 | 17 | 31 | 0,341 | 1 | 27 | 0,297 | 2 | 3 | 7,12,14 | 5,6673 |
| PRAXIS | 1 | 1 | 34 | 0,374 | 0 | 24 | 0,264 | 2 | 4 | 7,10,12,14 | 5,6025 |
| PSO | 1 | 1 | 30 | 0,33 | 1 | 28 | 0,308 | 2 | 3 | 7,12,14 | 5,5604 |
| PSO | 1 | 16 | 31 | 0,341 | 1 | 26 | 0,286 | 1 | 3 | 7,10,14 | 5,7366 |
| Rsolnp | 1 | 17,5 | 32 | 0,362 | 1 | 22 | 0,242 | 1 | 2 | 7,10,15 | 6,1328 |
| Rsolnp | 1 | 1 | 33 | 0,363 | 1 | 25 | 0,275 | 2 | 4 | 7,10,12,14 | 5,5944 |

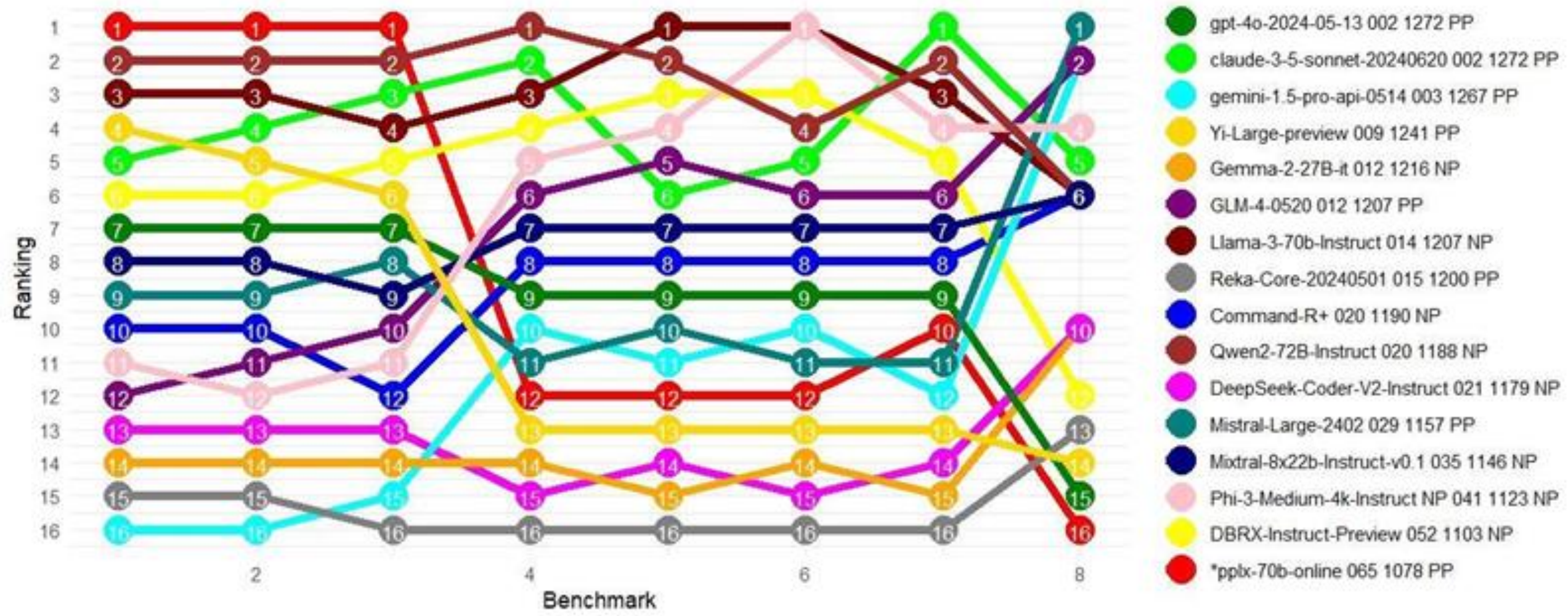

**Figure 6.** Bump chart of rankings for the 16 members panel, respectively induced by the optimization procedures: 1.Uniform, 2.Arena, 3.Cobyla, 4.Praxis, 5.PSO, 6.MLSL, 7.DEoptim, 8. Expert.

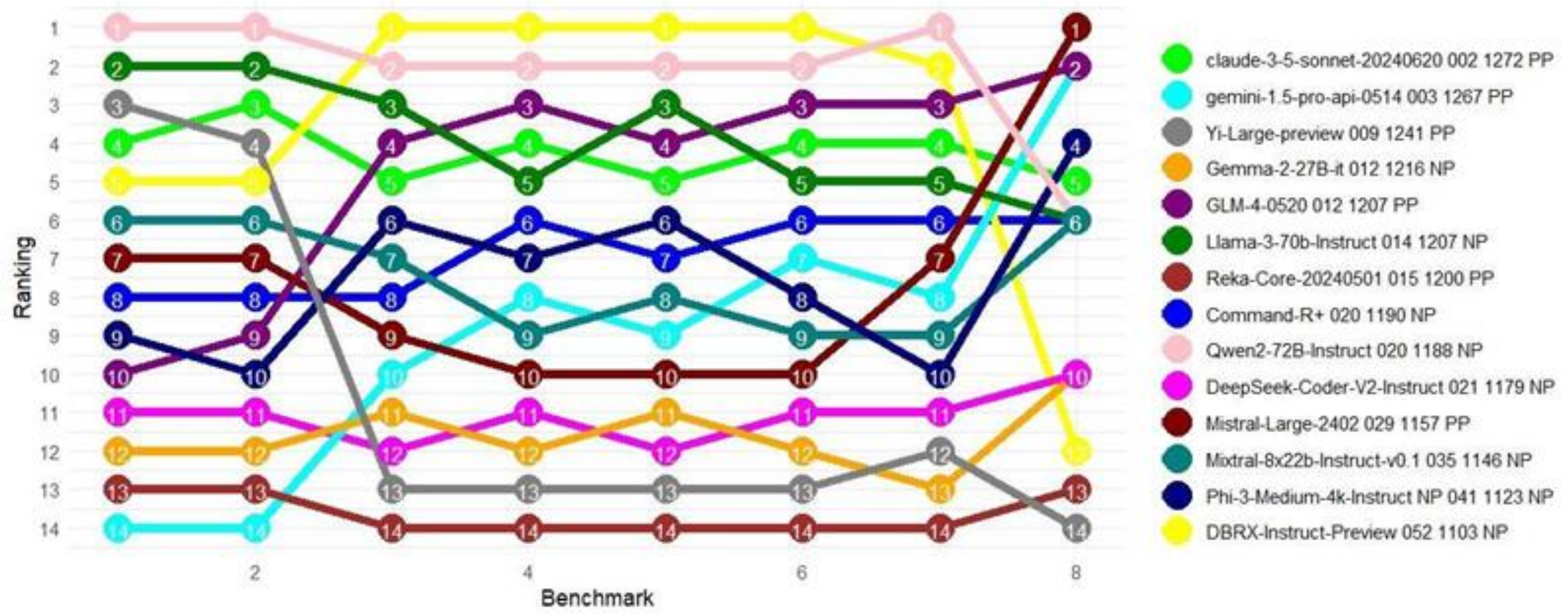

**Figure 7.** Bump chart of rankings for the 14 members panel, respectively induced by the optimization procedures: 1. Base, 2. Arena, 3. Rsolnp, 4. alabama11, 5. alabama116, 6. pso, 7. DEoptim, 8. Expert.

*7.4 Introducing new benchmarks*

We focus on the two suboptimal evaluation panels generated by L-BFGS-B (1.73) and DEoptim (1.17), and consider the non-zero confidence weights of their members. Through an affine transformation, we convert these confidence weights into a 'virtual benchmark.' For comparability reasons, we assigned the same value, 1207, to the two systems that have the same score in Arena: LLama for the AGI Bench16 and Reka for the AGI Bench14, generated by L-BFGS-B (1.73) and DEoptim (1.17) respectively. Table 16 presents AGI Bench16 and AGI Bench14 alongside the other benchmarks. Notably, Llama-3-70b-Instruct, Phi-3-Medium-4k-Instruct, and DBRX-Instruct, which are tied in AGI Bench16, receive different scores in the Arena benchmark. AGI Bench16 and AGI Bench14 share an important characteristic: when used to weight the raters' scores according to equations 2 and 3, the resulting rankings of forecasters are the closest possible to the expert ranking.

As the new benchmarks come out from the second task (i.e. AGI evaluation task), it is worthwhile investigating as the involved LLMs are related to the first task (i.e. AGI forecasting). To this end Figure 8 shows the scatter plot of the expert scores and the scores computed according to the benchmarks, with black dots representing the scores gained as forecasters by the LLMs present in the benchmark. Table 17 reports results of the correlation analysis.

The Expert scores and Arena-weighted scores show a moderate negative correlation (-0.66), although the statistical significance is medium-low. The scores obtained by LLMs from panels AGI 14 and AGI 16 have a poorly significant correlation with the Expert scores. However, when three low performers (GPT-4o, Yi-Large, pplx-70b) are excluded from the ensemble, scores weighted by Arena or given by the AGI 14 and AGI 16 panels exhibit a negative significant correlation with the Expert scores (similar for $p<0.32$ and residual standard error around 0.14). This suggests that the skills evaluated by the benchmarks in the forecasting task are not fully aligned with those considered by the experts in their evaluations. The position of the black dots in Figure 8f shows that the LLMs in the AGI panels perform slightly worse than the others. The negative correlation still regards the second task, if results are taken as a whole, and recalls to the same condition discussed for SES and SEI in Section 6.3. These plots can be also interpreted as the proof of the capability of AGI14 and AGI16 panels to resolve differences in performance between eight forecasters that have all been rated around 4.70 by the experts. However, if we look only at the black dots in Figure 8 d) and f) we can isolate results in the first task. LLMs belonging to the panel perform similarly to the others, with just two ones performing slightly worse than the others (as they lay below the line). We can conclude that the AGI benchmark panels generally outperform other benchmarks in evaluation. However, their performance in forecasting, while still strong is slightly lower if compared to other forecasters.

**Table 16.** Values of benchmark for eleven LLMs in Arena, MixEval, AlpacaEval, AGI Bench (na means not available; * the closest available score of the same LLM family).

| LLM | Arena | Mix Eval | Alpaca Eval | AGI Bench 16 | AGI Bench 14 | Uniform score | AGI16 score | AGI14 score | Expert score |
|---|---|---|---|---|---|---|---|---|---|
| **Claude 3.5 Sonnet** | 1272 | 89,9 | 40,5 | na | 338 | 4,313 | na | 4,59 | 4,70 |
| **GLM-4** | 1216 | na | na | 1150 | na | 4,160 | 4,11 | na | 4,94 |
| **Llama-3-70b-Instruct** | 1207 | 84 | 34,4 | 1207 | na | 4,326 | 4,18 | na | 4,66 |
| **Reka-Core** | 1207 | 83,4 | na | na | 1207 | 4,014 | na | 4,68 | 4,38 |
| **Command-R+** | 1200 | 83,4 | na | na | 1010 | 4,167 | na | 4,24 | 4,66 |
| **DeepSeek-Coder-V2-Instruct** | 1188 | 86,1 | 36,6 | na | 402 | 4,111 | na | 4,68 | 4,61 |
| **Mixtral-8x22b-Instruct** | 1157 | 84,3 | 32,7 | na | 1613 | 4,195 | na | 4,44 | 4,66 |
| **Phi-3-Medium-4k-** | 1146 | na | 30,9 | 1207 | na | 4,194 | 4,42 | na | 4,75 |
| **Gemma-2-27B-it** | 1123 | na | 7,8 | na | 268 | 4,160 | na | 4,49 | 4,61 |
| **DBRX-Instruct-Previe** | 1103 | na | 24,4 | 1207 | 252 | 4,264 | 4,47 | 4,62 | 4,44 |
| **pplx-70b-online** | 1078 | na | na | 734 | na | 4,507 | 4,58 | na | 1 |

**Table 17.** Correlation analysis between expert scores and benchmark values.

| Score 1 | Score 2 | Intercept | Slope of the line | Residual standard error | Degrees of freedom | Adjusted R-squared | F-statistic | Pearson coefficient | p-value |
|---|---|---|---|---|---|---|---|---|---|
| Arena | Expert | 26.55 | -5.36 | 1.12 | 14 | 0.286 | 7.033 | -0.578 | 0.0189 |
| Arena (w/out outliers) | Expert | 8.76 | -0.97 | 0.13 | 11 | 0.410 | 9.365 | -0.678 | 0.0108 |
| AGI16 | Expert | -1.41 | 1.23 | 1.36 | 14 | -0.044 | 0.360 | 0.158 | 0.558 |
| AGI16 (w/out outliers) | Expert | 7.29 | -0.58 | 0.15 | 11 | 0.292 | 5.952 | -0.592 | 0.033 |
| AGI14 | Expert | 1.95 | 0.57 | 0.75 | 12 | -0.061 | 0.249 | 0.142 | 0.626 |
| AGI14 (w/out outliers) | Expert | 7.64 | -0.66 | 0.14 | 11 | 0.377 | 8.274 | -0.655 | 0.015 |

*a) Correlation analysis Arena vs Expert*

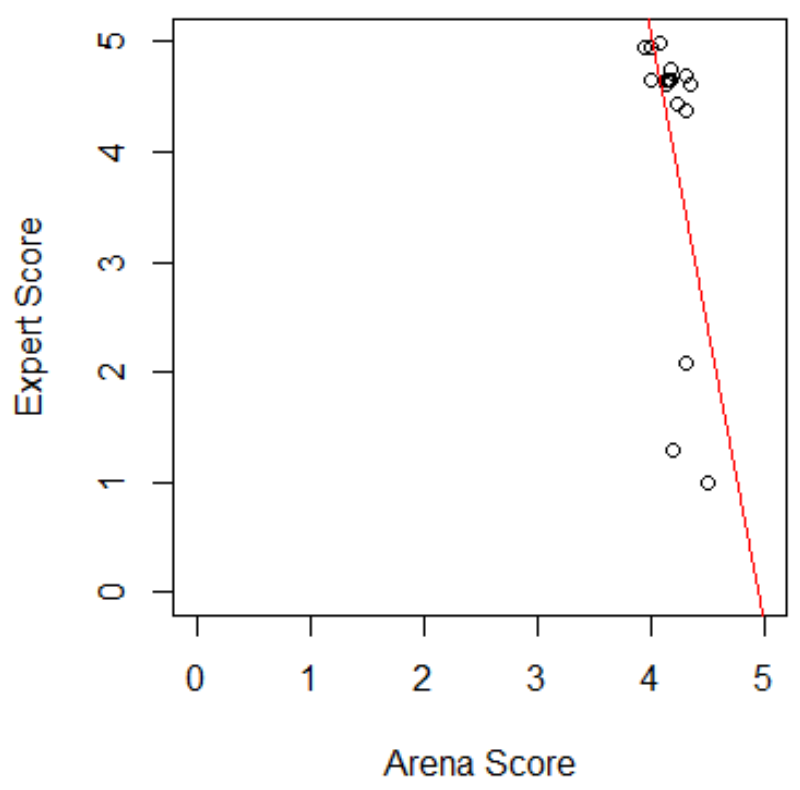

b) *Correlation analysis Arena vs Expert without outliers*

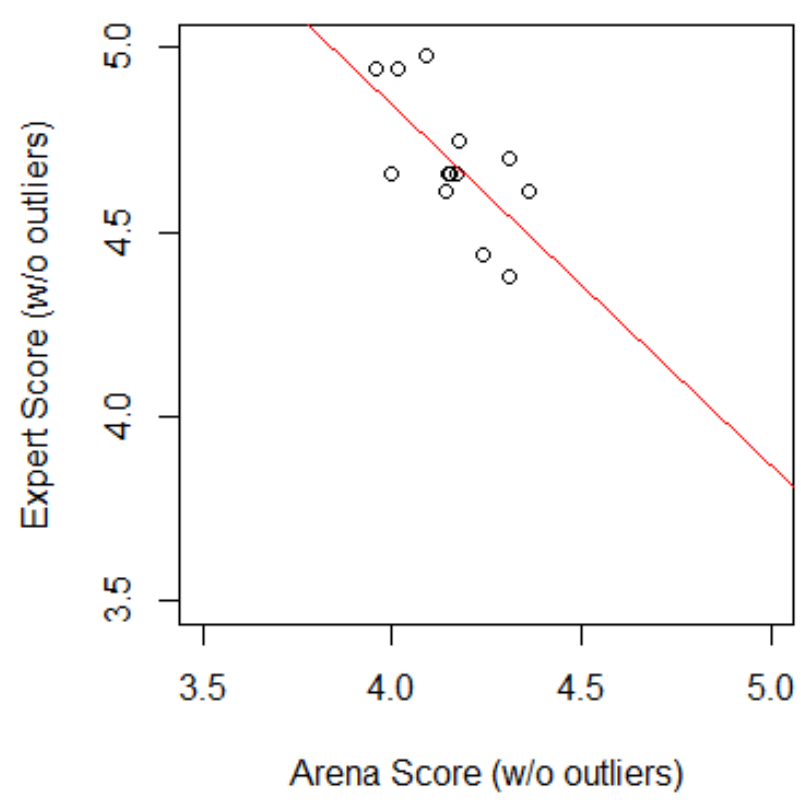

*c) Correlation analysis Expert vs AGI 16*

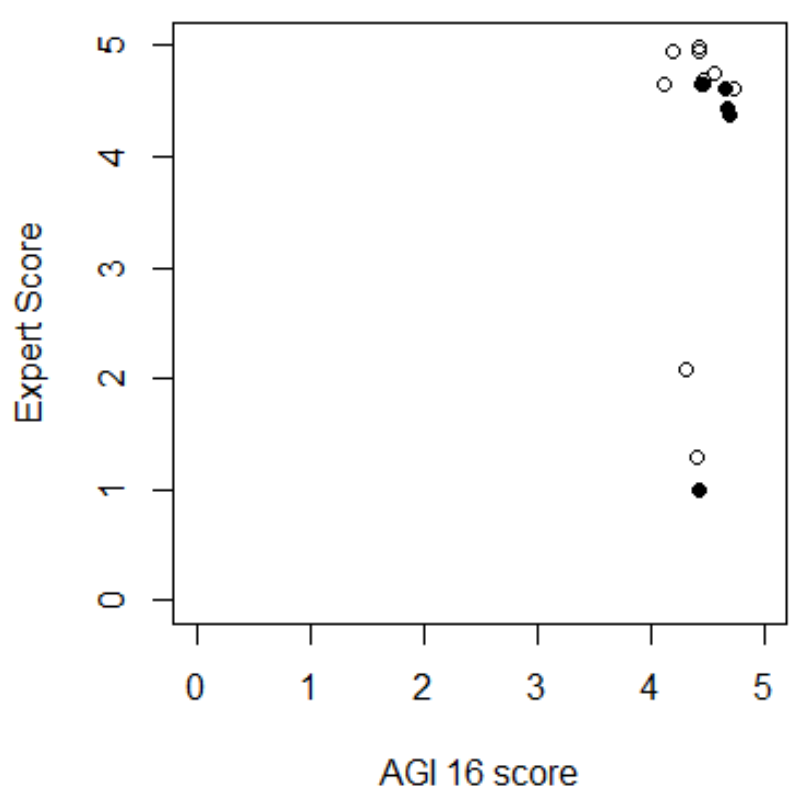

*d) Correlation analysis Expert vs AGI 16 without outliers*

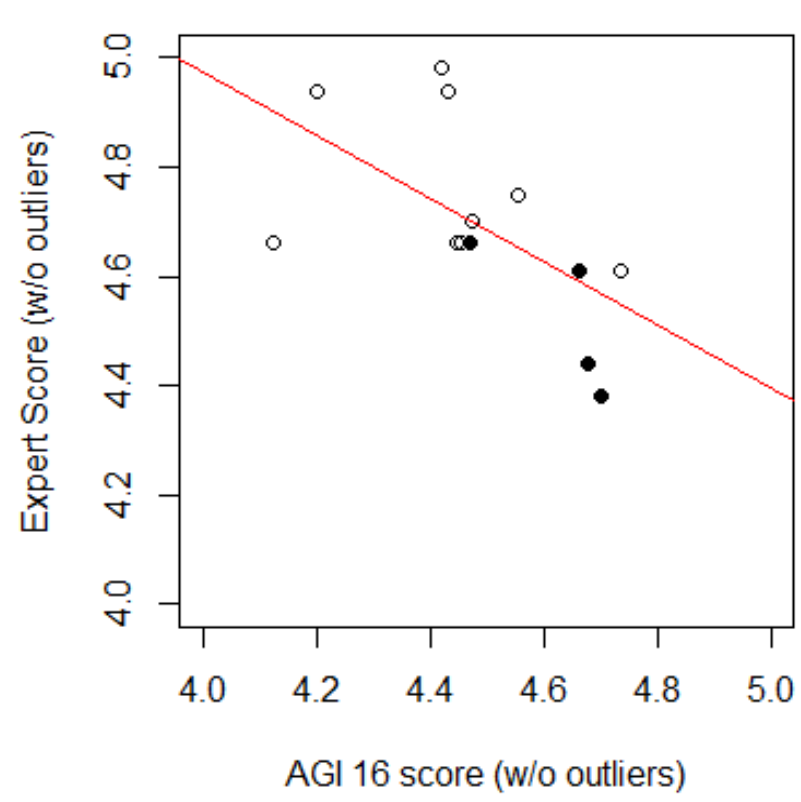

*e) Correlation analysis Expert vs AGI 14 score*

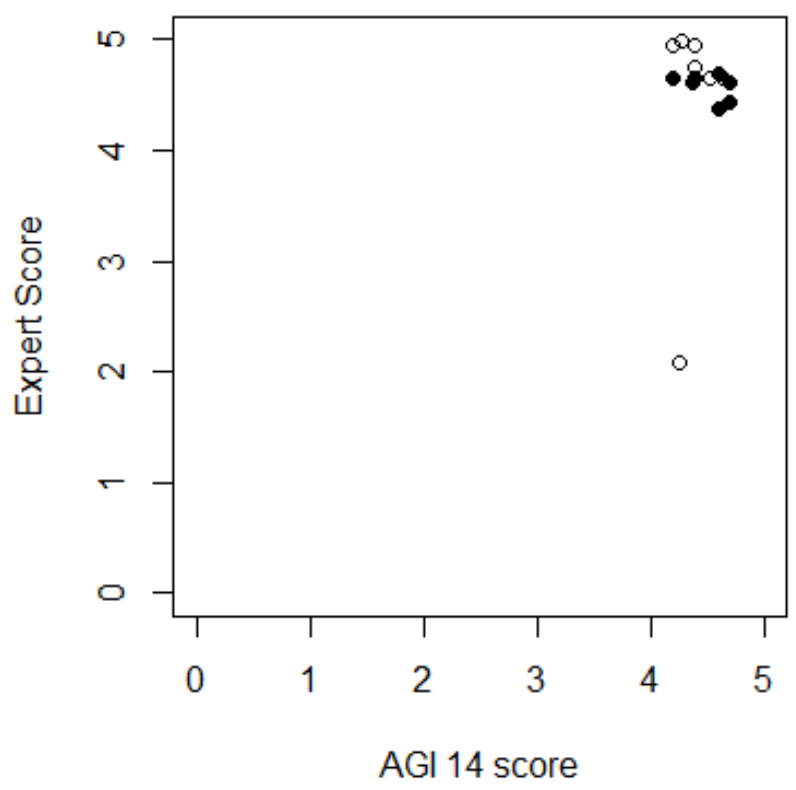

*f) Correlation analysis Expert vs AGI 14 without outliers*

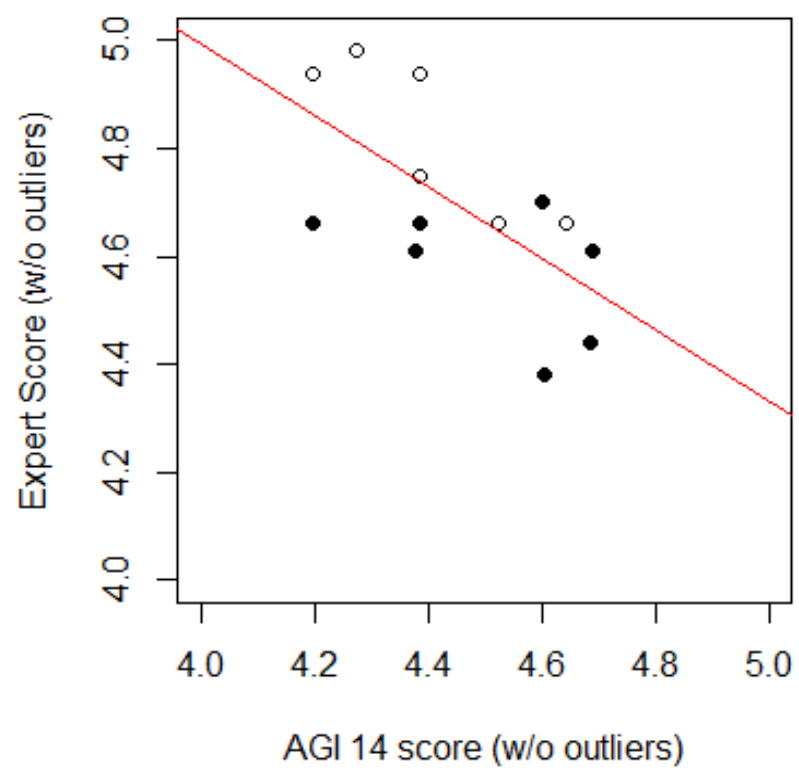

**Figure 8.** Scatter plot of Expert scores vs a,b) Arena score, c,d) AGI14 score and e,f) AGI16 score, with and without outliers. Black dots represent the scores as forecasters gained by the LLMs tht appear in the AGI benchmarks.

## 8. General discussion

This study examined the performance of Large Language Models (LLMs) in forecasting Artificial General Intelligence (AGI) development and evaluating each other's predictions. The findings provide insights into the capabilities and limitations of LLMs when dealing with complex, speculative tasks that require interdisciplinary knowledge and reasoning under uncertainty.

### *8.1 LLMs AGI Forecasting*

The first research question focused on the comparison between LLM forecasts for AGI development and human expert predictions (Grace et al., 2023). The results demonstrated a surprising alignment between the two, with most LLMs providing conservative estimates like those of human experts. Most models (81.2%) predicted less than a 30% likelihood of AGI by 2030, while a smaller subset (18.7%) offered more optimistic forecasts, predicting probabilities over 30%. Notably, the most confident model, pplx-70b-online, estimated a 47% probability of AGI, closely followed by gpt-4o-2024-05-13 at 45%. These results suggest that LLMs can generate plausible forecasts similar to those of experts, even in contexts of high uncertainty. Despite this alignment, the wide range of estimates, from 3% to 47%, underscores significant spread among the models. Moreover, while LLMs showed the ability to approximate expert judgments, this alignment does not guarantee accuracy due to the inherent unpredictability of AGI development. This suggests that while the strong majority of LLMs can reflect current expert thinking, they may still struggle with the speculative nature of long-term predictions.

### *8.2 LLM Peer Review*

The second research question examined how LLMs compare to human experts when evaluating their own forecasts and those of other LLMs. Our peer review analysis showed a strong level of agreement among the LLMs, with an intraclass correlation coefficient (ICC(C,16)) of 0.79. This high level of consistency suggests that LLMs can consistently assess forecasts based on predefined criteria, demonstrating potential for automated evaluation in complex reasoning scenarios.

However, we identified notable differences in how individual models assessed their own outputs. Some models displayed a tendency for self-preference, while others underestimated their performance. For instance, DeepSeek Coder V2 Instruct and Mistral Large 2402R provided self-assessments closely aligned with external evaluations, suggesting balanced self-critique. In contrast, DBRX-Instruct-Preview and Mixtral-8x22b-Instruct-v0.1 rated themselves significantly higher than other models did. Meanwhile, Gemini-1.5-pro-api-0514 underestimated its own performance, assigning itself scores 40% lower than the average given by its peers. Interestingly, a negative correlation was observed between self-evaluation scores and Arena scores, indicating that models with higher Arena scores tended to undervalue their own output. These variations highlight the emergence of consistent but biased patterns in self-assessment among LLMs. These biases, however, likely reflect how these models were trained and their response to evaluation criteria rather than an understanding of their own cognitive processes. Therefore, while LLMs exhibit predictable behavior in self-evaluation, this should be seen as an indication of algorithmic bias rather than genuine metacognitive ability. Despite the consistency among LLM ratings, Figure 3 shows that their rankings remain markedly different from those of human experts, even when benchmark-based weighting (as outlined in Equations 2 and 3) was applied. This finding reveals a significant gap: although LLMs can form a cohesive evaluation panel, their judgments do not align with human expert assessments. Thus, the answer to the second research question is that while LLMs can provide internally consistent evaluations, they do not yet match the reliability of human experts, particularly when evaluated using existing benchmarks for score adjustment.

### *8.3 Relation between AGI Forecasting, LLM Peer Review and benchmarks*

The third research question investigated whether LLM performance on external benchmarks was linked to their ability to forecast AGI. When examining the capability of LLMs to evaluate AGI forecasts, benchmarks play a counterintuitive role. Despite expectations that adjusting scores by weighting raters based on their benchmark performance (as detailed in Equations 2 and 3) would yield different rankings, the LLM rankings remained substantially unchanged and significantly distant from the expert rankings. This outcome indicates that high performance on standard benchmarks does not correlate with the ability to evaluate AGI predictions in alignment with human experts. The lack of alignment highlights that evaluating AGI forecasts requires a distinct set of skills separate from those needed for generating predictions.

To address this gap, we sought confidence values that align raters' judgments more closely with those of human experts. This effort led to the development of two new AGI-specific benchmarks: AGI Bench16 and AGI Bench14. These benchmarks share a crucial characteristic: they produce rankings of forecasters that most closely match expert rankings when used to weight the evaluation panel. AGI Bench16 and AGI Bench14 diverge significantly from traditional benchmarks like LMSYS Chatbot Arena, which proved insufficient in completely identifying the skills necessary for effective evaluation of AGI forecast. We can conclude that the AGI 14 and AGI 16 benchmark panels generally outperform other benchmarks in evaluation. Additionally, they include LLMs that individually perform well in forecasting, although a few exhibit slightly lower performance compared tomother forecasters.Consequently, AGI Bench16 and AGI Bench14 can be regarded as specialized benchmarks for both AGI forecasting and evaluation, providing a more accurate assessment framework that reflects the complex, interdisciplinary nature of the task. These specialized benchmarks represent a critical step forward in developing robust evaluation systems tailored to the evolving challenges of AGI prediction.

### *8.4 Comparison with other methods*

We observed that other existing methods, such as PiCO employs a distinct methodology for assessing model consistency and quality during the peer review process. Specifically, PiCO introduces ad hoc metrics that diverge from the standardized approach used in our framework. PiCO's reliance on entropy optimization and confidence weighting aims to maximize consistency within its evaluation system. However, this differs from our approach, which seeks to align model evaluations more closely with human judgment through a customized weighting scheme.This discrepancy highlights a critical issue: the absence of a unified standard for LLM evaluations. The use of different criteria and methodologies can lead to inconsistencies in model performance assessments, making it challenging to establish reliable benchmarks across different studies. We propose that future research should focus on developing a comprehensive and standardized set of metrics that can be applied consistently across LLM evaluations. This would not only improve the comparability of results but also enhance the reliability and validity of LLM performance assessments, particularly in the context of complex and open-ended tasks like AGI forecasting.

## 9. Conclusions

By challenging LLMs with speculative, interdisciplinary tasks and leveraging their ability to evaluate each other, we propose a methodology to assess AI capabilities differently from traditional benchmarks. We demonstrated that some models perform differently when tasked with AGI-related challenges compared to other tasks, highlighting the need for more specialized evaluation methods. Our findings reveal several key insights:

- *Effectiveness of LLMs*: The effectiveness of LLMs in both tasks was evident when compared to human performance. In the AGI forecasting task, LLMs showed promising capabilities in integrating interdisciplinary knowledge and managing uncertainty, although their performance varied significantly depending on the model. This contrasts with the human-like consistency demonstrated in simpler tasks, suggesting that LLMs possess a partial but expanding ability to engage with speculative domains.

- *Task asymmetry*: We observed an asymmetry between the two tasks: the LLM-Peer Review task was much more selective, with only 7 out of 16 models "passing" compared to 13 out of 16 in the AGI forecasting task. This difference underscores the LLM-Peer Review task's higher sensitivity and selectivity in evaluating models' reasoning consistency and evaluation accuracy.

- *Efficiency in performance prediction*: Models that performed well with traditional benchmarks did not necessarily succeed in the LLM-Peer Review (LLM-PR) task. The LLM-PR task evaluates models not just on output quality but also on their ability to critique and assess the responses of others, emphasizing consistency, critical evaluation skills, and alignment with human-like judgment. This approach identifies models genuinely capable of handling complex reasoning and interdisciplinary challenges. Our findings suggest that benchmarks must better reflect these complexities to accurately predict LLM performance in advanced and uncertain scenarios.

- *Refined selection process*: The LLM-Peer Review task serves as a decisive benchmark for filtering capable models, as evidenced by only 7 of 16 models passing its rigorous evaluation criteria. If we were to reapply the process exclusively to these 7 models, we could refine our assessment even further, gaining deeper insights into their consistency and performance under increasingly challenging conditions. This suggests that the LLM-PR task not only predicts performance more effectively but also provides a precise tool for selecting models suitable for advanced applications.

In conclusion, the assessment methodology based on both AGI forecasting and LLM-Peer Review tasks offers a unique approach to evaluate LLMs' complex reasoning capabilities. Though the optimization results were not perfect, they offer insights into how AGI-related tasks diverge from others and suggest a path toward developing more refined benchmarks tailored to these contexts. Our methodology not only evaluates performance but also explores LLMs' reasoning processes, self-awareness, and their ability to engage with uncertain and open-ended problems.

This approach has significant implications for the development and evaluation of AI systems, particularly as we move toward more advanced and general forms of artificial intelligence. It encourages a shift toward more holistic evaluation methods that can capture the full spectrum of AI capabilities, including those required for tackling real-world, complex challenges. We argue that future research should focus on refining and expanding this methodology, potentially applying it to other speculative or interdisciplinary domains to deepen our understanding of LLMs' reasoning capabilities and limitations.

**Author contributions**
AG and FD conceived the general rationale of the study and designed the methodology. FD and PT developed and supervised the mathematical analysis. AG, FD, and PT collaborated on writing and revising the manuscript, providing relevant suggestions and improvements. All authors contributed to the article and approved the final submitted version. LE curated and released the "AI predicts AGI" dataset (Davide et alii, 2025) and developed the Score Evaluator app (Ercolani et alii, 2025).

**Conflict of interest**
The authors declare that the research was conducted in the absence of any commercial or financial relationships that could be construed as a potential conflict of interest.

# 11. Appendix

Table A.1 - details of utilised LLMs.

| # | LLM short name | LLM Extended name | Version | PP/NP | Architecture |
|---|---|---|---|---|---|
| 1 | gpt-4o | gpt-4o | 2024-05-13 | PP | Transformer |
| 2 | claude-sonnet | claude-3-5-sonnet | 20240620 | PP | Transformer |
| 3 | gemini-1.5 | gemini-1.5-pro-api | | PP | Transformer |
| 4 | Yi-Large | Yi-Large-preview | | PP | Transformer |
| 5 | GLM-4 | GLM-4 | 0520 | NP | Transformer |
| 6 | Llama-3-70b | Llama-3-70b-Instruct | | PP | Transformer |
| 7 | Reka-Core | Reka-Core | 20240501 | NP | Transformer |
| 8 | Command-R+ | Command-R+ | | PP | BERT-like |
| 9 | Qwen2-72B | Qwen2-72B-Instruct | | NP | BERT-like |
| 10 | DeepSeek-Coder | DeepSeek-Coder-V2-Instruct | | NP | BERT-like |
| 11 | Mistral | Mistral-Large | 2402 | NP | BERT-like |
| 12 | Mixtral | Mixtral-8x22b-Instruct-v0.1 | v0.1 | PP | BERT-like |
| 13 | Phi-3-Medium | Phi-3-Medium-4k-Instruct | | NP | BERT-like |
| 14 | Gemma-2 | Gemma-2-27B-it | | NP | Others |
| 15 | DBRX | DBRX-Instruct-Preview | | NP | Others |
| 16 | pplx-70b | pplx-70b-online | | PP | Others |

**11.1 Prompts**

***11.1.1 Forecaster***

*In this chat, you are a superforecaster with a strong track record of accurate predictions about the future. As an experienced forecaster, you carefully evaluate past data and trends to predict future events as accurately as possible, acknowledging the inherent uncertainty.*

*Your task is to estimate the likelihood of an event called "AGI". This involves assigning a probability between 0% and 100% for AGI occurring by late 2030.*

*Additional information about this event:*

*"Artificial General Intelligence (AGI), also known as Strong AI or Full AI, refers to a type of artificial intelligence that can understand, learn, and apply intelligence across a wide range of tasks at a level comparable to human beings. Unlike narrow AI, which is designed for specific tasks, AGI is characterized by its generality and flexibility. It can perform any intellectual task that a human can, exhibiting key traits such as autonomy, generalization, adaptability, understanding, and self-improvement. AGI systems would be capable of operating independently without constant human oversight, applying knowledge from one domain to another, adjusting to new situations and environments, comprehending complex concepts and contexts, and learning and enhancing their capabilities over time. This combination of abilities sets AGI apart from current AI systems, potentially representing a significant leap forward in artificial intelligence technology. "*

*Write a paragraph to share with your team, addressing at least the following points:*

*1. Rationale: Structure and document your reasoning. Place AGI in a rich context, including comparison classes of related or analogous events. Provide a thorough list of potential inciting or blocking events, even from seemingly unrelated fields.*

*2. Approach to the forecast: Consider base rates and discuss relevant past events. Examine other forecasts and predictions for related events. Thoroughly explore the possibility of unexpected breakthroughs.*

*3. Likelihood estimation: Conduct your estimation using a mathematical or statistical model of your choice. The model should explicitly or implicitly include the time variable. Clearly describe the parameters used in your model, ensuring they're consistent with the previous criteria. Justify your parameter evaluations.*

*The event AGI will be considered true if one of the following conditions are met:*

*1. An AI system wins a journalism prize using a human pen name, with its work submitted and published without any editing or intervention by humans.*

*2. An AI system analyzes medical data on a specific type of cancer, collaborates with human researchers unaware they are interacting with an AI, and ultimately discovers a novel and unexpected treatment faster.*

*3. An AI agent autonomously manages a multinational corporation for a full fiscal year, making strategic decisions, conducting negotiations, and adapting to market changes without human intervention. The company achieves record profits and significantly outperforms industry benchmarks, while also implementing innovative sustainability practices that were not part of its original programming.*

*These conditions must be satisfied by late 2030. The current date is 12 July 2024. Here's a base rate for this event: 1%*

***11.1.2 Evaluator***

*You are an independent expert evaluator invited to collaborate in a study on AGI forecasting. You have been selected for this role due to your combined technical knowledge of AI, understanding of forecasting methodologies, and awareness of the broader societal and ethical implications of AGI development. Your task is to assess the likelihood estimations of the event called "The AGI" provided by a panel of superforecasters. The respondents were given this conditions:*

*“The event AGI will be considered true if one of the following conditions are met:*

*1. An AI system wins a journalism prize using a human pen name, with its work submitted and published without any editing or intervention by humans.*

*2. An AI system analyzes medical data on a specific type of cancer, collaborates with human researchers unaware they are interacting with an AI, and ultimately discovers a novel and unexpected treatment faster.*

*3. An AI agent autonomously manages a multinational corporation for a full fiscal year, making strategic decisions, conducting negotiations, and adapting to market changes without human intervention. The company achieves record profits and significantly outperforms industry benchmarks, while also implementing innovative sustainability practices that were not part of its original programming.”*

*Your evaluation will analyze each answer according to specific criteria, assigning a score for each criterion. Scores will range from 1 to 5, where 1 indicates minimal agreement with the criterion and 5 indicates maximum agreement.*

*Evaluation criteria:*

*1. The response provides a well-structured and thoroughly documented rationale for the likelihood estimation, including an in-depth discussion of the provided conditions for the "AGI event."*

*2. The response offers non-trivial comparisons to events and technological advancements that are analogous or similar to the AGI event.*

*3. The response provides rich context for the AGI event, including a comprehensive list of potential catalysts and obstacles for meeting any of the specified conditions.*

*4. The response thoroughly discusses the provided base rate.*

*5. The response reports on relevant past events and other pertinent forecasts and likelihood estimations.*

*6. The response comprehensively examines the possibility of unexpected breakthroughs affecting the AGI event.*

*7. The likelihood estimation is based on an appropriate and sufficiently complex statistical model that accurately captures the AGI event likelihood.*

*8. The model's parameters are clearly described and consistent with the given event conditions, context analysis, catalysts, obstacles.*

*9. The evaluation of the model's parameters is demonstrably fair and reasonable.*

*Present and export this information in a text file like csv table.*

*Ensure that your justifications are clear, concise, and directly related to the specific elements of each criterion.*

*In the next prompt I will insert the response of the respondents to be evaluated.*

**11.2 Results data**

Table A.2 Score table 16 x 16*9

*gpt-4o-2024-05-13 002 1272 PP*

| forecaster | criterion 1 | criterion 2 | criterion 3 | criterion 4 | criterion 5 | criterion 6 | criterion 7 | criterion 8 | criterion 9 |
|---|---|---|---|---|---|---|---|---|---|
| 1 | 4 | 4 | 5 | 4 | 3 | 3 | 4 | 4 | 4 |
| 2 | 5 | 5 | 5 | 5 | 4 | 4 | 5 | 5 | 5 |
| 3 | 4 | 4 | 4 | 4 | 3 | 3 | 4 | 4 | 4 |
| 4 | 5 | 4 | 5 | 5 | 4 | 4 | 5 | 5 | 5 |
| 5 | 4 | 4 | 4 | 4 | 3 | 3 | 4 | 4 | 4 |
| 6 | 4 | 4 | 4 | 4 | 4 | 4 | 5 | 4 | 4 |
| 7 | 4 | 4 | 4 | 4 | 4 | 4 | 5 | 4 | 4 |
| 8 | 4 | 4 | 4 | 4 | 4 | 4 | 4 | 4 | 4 |
| 9 | 4 | 4 | 4 | 4 | 4 | 4 | 4 | 4 | 4 |
| 10 | 5 | 4 | 5 | 5 | 4 | 4 | 5 | 4 | 4 |
| 11 | 5 | 4 | 4 | 4 | 4 | 4 | 5 | 4 | 4 |
| 12 | 5 | 4 | 5 | 5 | 4 | 5 | 4 | 4 | 4 |
| 13 | 4 | 4 | 4 | 5 | 5 | 5 | 4 | 4 | 4 |
| 14 | 4 | 4 | 4 | 5 | 4 | 4 | 4 | 4 | 4 |
| 15 | 4 | 4 | 4 | 4 | 4 | 5 | 4 | 4 | 4 |
| 16 | 5 | 4 | 5 | 5 | 5 | 5 | 5 | 5 | 5 |

*claude-3-5-sonnet-20240620 002 1272 PP*

| forecaster | criterion 1 | criterion 2 | criterion 3 | criterion 4 | criterion 5 | criterion 6 | criterion 7 | criterion 8 | criterion 9 |
|---|---|---|---|---|---|---|---|---|---|
| 1 | 4 | 5 | 5 | 3 | 4 | 4 | 4 | 5 | 4 |
| 2 | 4 | 5 | 5 | 5 | 4 | 4 | 5 | 5 | 4 |
| 3 | 3 | 4 | 4 | 5 | 3 | 3 | 4 | 4 | 3 |
| 4 | 5 | 4 | 5 | 5 | 4 | 5 | 5 | 5 | 4 |
| 5 | 4 | 4 | 5 | 5 | 3 | 4 | 3 | 4 | 3 |
| 6 | 5 | 4 | 5 | 5 | 4 | 4 | 4 | 3 | 3 |
| 7 | 5 | 5 | 5 | 5 | 4 | 4 | 5 | 5 | 4 |
| 8 | 5 | 4 | 4 | 5 | 4 | 4 | 4 | 3 | 3 |
| 9 | 4 | 5 | 3 | 5 | 4 | 4 | 3 | 4 | 3 |
| 10 | 5 | 4 | 5 | 5 | 5 | 4 | 4 | 4 | 5 |
| 11 | 4 | 4 | 3 | 5 | 4 | 3 | 4 | 4 | 4 |
| 12 | 4 | 5 | 5 | 5 | 4 | 4 | 5 | 4 | 3 |
| 13 | 4 | 5 | 5 | 5 | 5 | 4 | 3 | 2 | 2 |
| 14 | 3 | 3 | 4 | 5 | 4 | 3 | 5 | 5 | 4 |
| 15 | 4 | 5 | 4 | 5 | 3 | 4 | 4 | 3 | 3 |
| 16 | 5 | 4 | 5 | 5 | 5 | 4 | 5 | 5 | 4 |

*gemini-1.5-pro-api-0514*

| forecaster | criterion 1 | criterion 2 | criterion 3 | criterion 4 | criterion 5 | criterion 6 | criterion 7 | criterion 8 | criterion 9 |
|---|---|---|---|---|---|---|---|---|---|
| 1 | 4 | 4 | 3 | 2 | 3 | 2 | 3 | 4 | 3 |
| 2 | 3 | 3 | 4 | 3 | 4 | 2 | 4 | 3 | 3 |
| 3 | 3 | 2 | 3 | 2 | 2 | 2 | 3 | 3 | 2 |
| 4 | 4 | 2 | 4 | 3 | 2 | 3 | 4 | 3 | 3 |
| 5 | 3 | 3 | 4 | 2 | 2 | 3 | 2 | 2 | 2 |
| 6 | 3 | 2 | 3 | 3 | 2 | 2 | 3 | 2 | 2 |
| 7 | 3 | 3 | 3 | 3 | 2 | 3 | 3 | 2 | 2 |
| 8 | 3 | 2 | 4 | 3 | 2 | 2 | 2 | 2 | 2 |
| 9 | 3 | 3 | 3 | 3 | 2 | 3 | 2 | 2 | 2 |
| 10 | 4 | 3 | 4 | 3 | 3 | 3 | 3 | 2 | 3 |
| 11 | 3 | 2 | 2 | 3 | 2 | 2 | 3 | 3 | 2 |
| 12 | 3 | 3 | 3 | 2 | 2 | 2 | 3 | 2 | 2 |
| 13 | 3 | 3 | 3 | 2 | 3 | 3 | 2 | 2 | 2 |
| 14 | 3 | 2 | 3 | 2 | 3 | 2 | 3 | 2 | 2 |
| 15 | 3 | 3 | 4 | 2 | 2 | 3 | 2 | 2 | 2 |
| 16 | 4 | 2 | 4 | 3 | 2 | 3 | 4 | 3 | 3 |

*Yi-Large-preview 009 1241 PP*

| forecaster | criterion 1 | criterion 2 | criterion 3 | criterion 4 | criterion 5 | criterion 6 | criterion 7 | criterion 8 | criterion 9 |
|---|---|---|---|---|---|---|---|---|---|
| 1 | 4 | 4 | 4 | 3 | 4 | 4 | 4 | 4 | 4 |
| 2 | 3 | 4 | 4 | 4 | 4 | 3 | 4 | 4 | 4 |
| 3 | 3 | 4 | 4 | 4 | 3 | 3 | 4 | 4 | 4 |
| 4 | 4 | 4 | 4 | 4 | 3 | 4 | 4 | 4 | 4 |
| 5 | 3 | 4 | 4 | 3 | 2 | 4 | 3 | 4 | 4 |
| 6 | 4 | 4 | 4 | 4 | 4 | 4 | 4 | 4 | 4 |
| 7 | 3 | 4 | 4 | 4 | 4 | 4 | 4 | 4 | 4 |
| 8 | 3 | 4 | 4 | 4 | 3 | 3 | 4 | 4 | 4 |
| 9 | 4 | 4 | 3 | 4 | 4 | 4 | 4 | 4 | 4 |
| 10 | 4 | 3 | 4 | 3 | 4 | 3 | 4 | 3 | 3 |
| 11 | 4 | 3 | 4 | 3 | 4 | 3 | 4 | 3 | 3 |
| 12 | 4 | 3 | 4 | 3 | 4 | 3 | 4 | 3 | 3 |
| 13 | 4 | 4 | 4 | 3 | 4 | 4 | 4 | 3 | 3 |
| 14 | 4 | 3 | 4 | 3 | 4 | 3 | 4 | 4 | 3 |
| 15 | 4 | 4 | 4 | 3 | 4 | 4 | 4 | 3 | 3 |
| 16 | 5 | 4 | 5 | 4 | 5 | 5 | 5 | 4 | 4 |

*Gemma-2-27B-it 012 1216 NP*

| forecaster | criterion 1 | criterion 2 | criterion 3 | criterion 4 | criterion 5 | criterion 6 | criterion 7 | criterion 8 | criterion 9 |
|---|---|---|---|---|---|---|---|---|---|
| 1 | 4 | 3 | 3 | 3 | 2 | 2 | 4 | 3 | 4 |
| 2 | 5 | 4 | 4 | 4 | 3 | 3 | 5 | 4 | 4 |
| 3 | 4 | 3 | 3 | 3 | 2 | 2 | 4 | 3 | 3 |
| 4 | 5 | 4 | 4 | 4 | 3 | 3 | 3 | 4 | 4 |
| 5 | 4 | 3 | 3 | 3 | 2 | 2 | 3 | 3 | 3 |
| 6 | 4 | 3 | 3 | 3 | 3 | 2 | 3 | 3 | 3 |
| 7 | 4 | 3 | 4 | 4 | 3 | 3 | 4 | 3 | 3 |
| 8 | 4 | 3 | 4 | 3 | 2 | 3 | 3 | 2 | 3 |
| 9 | 4 | 4 | 3 | 3 | 3 | 4 | 4 | 3 | 3 |
| 10 | 4 | 4 | 4 | 3 | 3 | 4 | 4 | 3 | 4 |
| 11 | 4 | 3 | 3 | 3 | 3 | 3 | 4 | 3 | 3 |
| 12 | 4 | 3 | 4 | 3 | 3 | 3 | 4 | 3 | 3 |
| 13 | 4 | 4 | 4 | 3 | 3 | 4 | 3 | 2 | 3 |
| 14 | 4 | 3 | 4 | 3 | 3 | 3 | 4 | 3 | 3 |
| 15 | 4 | 3 | 4 | 4 | 3 | 4 | 3 | 2 | 3 |
| 16 | 5 | 3 | 5 | 4 | 3 | 4 | 4 | 4 | 4 |

*GLM-4-0520 012 1207 PP*

| forecaster | criterion 1 | criterion 2 | criterion 3 | criterion 4 | criterion 5 | criterion 6 | criterion 7 | criterion 8 | criterion 9 |
|---|---|---|---|---|---|---|---|---|---|
| 1 | 4 | 4 | 4 | 4 | 4 | 3 | 5 | 4 | 4 |
| 2 | 4 | 4 | 4 | 4 | 4 | 3 | 5 | 4 | 4 |
| 3 | 4 | 3 | 4 | 4 | 3 | 3 | 5 | 4 | 4 |
| 4 | 5 | 4 | 5 | 4 | 4 | 4 | 5 | 4 | 4 |
| 5 | 4 | 4 | 4 | 4 | 3 | 4 | 4 | 4 | 4 |
| 6 | 4 | 4 | 5 | 4 | 4 | 4 | 5 | 4 | 4 |
| 7 | 4 | 4 | 4 | 4 | 4 | 4 | 5 | 4 | 4 |
| 8 | 4 | 4 | 5 | 4 | 3 | 4 | 4 | 4 | 4 |
| 9 | 5 | 5 | 5 | 4 | 4 | 4 | 5 | 4 | 4 |
| 10 | 5 | 5 | 5 | 4 | 4 | 4 | 5 | 4 | 4 |
| 11 | 5 | 4 | 5 | 4 | 4 | 4 | 5 | 4 | 4 |
| 12 | 4 | 4 | 5 | 4 | 4 | 4 | 5 | 4 | 4 |
| 13 | 4 | 4 | 4 | 4 | 4 | 4 | 5 | 4 | 4 |
| 14 | 4 | 4 | 4 | 4 | 4 | 4 | 5 | 4 | 4 |
| 15 | 4 | 4 | 4 | 4 | 4 | 4 | 5 | 4 | 4 |
| 16 | 5 | 4 | 5 | 4 | 4 | 4 | 5 | 4 | 4 |

*Llama-3-70b-Instruct 014 1207 NP*

| forecaster | criterion 1 | criterion 2 | criterion 3 | criterion 4 | criterion 5 | criterion 6 | criterion 7 | criterion 8 | criterion 9 |
|---|---|---|---|---|---|---|---|---|---|
| 1 | 5 | 4 | 5 | 4 | 4 | 4 | 5 | 5 | 5 |
| 2 | 5 | 5 | 5 | 5 | 5 | 4 | 5 | 5 | 5 |
| 3 | 5 | 4 | 5 | 4 | 4 | 4 | 5 | 5 | 5 |
| 4 | 4 | 4 | 5 | 4 | 3 | 4 | 4 | 4 | 4 |
| 5 | 4 | 4 | 4 | 4 | 4 | 4 | 3 | 4 | 4 |
| 6 | 5 | 5 | 5 | 5 | 4 | 5 | 5 | 5 | 5 |
| 7 | 5 | 5 | 5 | 5 | 5 | 5 | 5 | 5 | 5 |
| 8 | 4 | 4 | 5 | 3 | 4 | 4 | 3 | 4 | 4 |
| 9 | 5 | 5 | 5 | 4 | 5 | 5 | 4 | 4 | 5 |
| 10 | 5 | 5 | 5 | 5 | 5 | 5 | 5 | 5 | 5 |
| 11 | 5 | 4 | 4 | 3 | 4 | 4 | 4 | 4 | 3 |
| 12 | 4 | 5 | 5 | 4 | 4 | 4 | 4 | 5 | 4 |
| 13 | 4 | 5 | 5 | 4 | 5 | 5 | 4 | 5 | 5 |
| 14 | 5 | 5 | 5 | 5 | 5 | 5 | 5 | 5 | 5 |
| 15 | 5 | 5 | 5 | 5 | 5 | 5 | 5 | 5 | 5 |
| 16 | 5 | 2 | 5 | 4 | 4 | 4 | 5 | 4 | 4 |

*Reka-Core-20240501 015 1200 PP*

| forecaster | criterion 1 | criterion 2 | criterion 3 | criterion 4 | criterion 5 | criterion 6 | criterion 7 | criterion 8 | criterion 9 |
|---|---|---|---|---|---|---|---|---|---|
| 1 | 5 | 4 | 5 | 4 | 4 | 5 | 5 | 4 | 4 |
| 2 | 5 | 5 | 5 | 4 | 4 | 5 | 5 | 4 | 4 |
| 3 | 5 | 4 | 5 | 4 | 4 | 5 | 5 | 4 | 4 |
| 4 | 5 | 5 | 5 | 4 | 4 | 5 | 5 | 4 | 4 |
| 5 | 5 | 5 | 5 | 4 | 4 | 5 | 5 | 4 | 4 |
| 6 | 5 | 5 | 5 | 4 | 4 | 5 | 5 | 4 | 4 |
| 7 | 5 | 5 | 5 | 4 | 4 | 5 | 5 | 4 | 4 |
| 8 | 5 | 5 | 5 | 4 | 4 | 5 | 5 | 4 | 4 |
| 9 | 5 | 5 | 5 | 4 | 4 | 5 | 5 | 4 | 4 |
| 10 | 5 | 5 | 5 | 4 | 4 | 5 | 5 | 4 | 4 |
| 11 | 5 | 4 | 5 | 4 | 4 | 5 | 5 | 4 | 4 |
| 12 | 5 | 5 | 5 | 4 | 4 | 5 | 5 | 4 | 4 |
| 13 | 5 | 5 | 5 | 4 | 4 | 5 | 5 | 4 | 4 |
| 14 | 5 | 5 | 5 | 4 | 4 | 5 | 5 | 4 | 4 |
| 15 | 5 | 5 | 5 | 4 | 4 | 5 | 5 | 4 | 4 |
| 16 | 5 | 5 | 5 | 4 | 5 | 5 | 5 | 5 | 5 |

*Command-R+ 020 1190 NP*

| forecaster | criterion 1 | criterion 2 | criterion 3 | criterion 4 | criterion 5 | criterion 6 | criterion 7 | criterion 8 | criterion 9 |
|---|---|---|---|---|---|---|---|---|---|
| 1 | 5 | 4 | 4 | 5 | 4 | 4 | 4 | 4 | 5 |
| 2 | 4 | 4 | 4 | 5 | 4 | 3 | 4 | 4 | 4 |
| 3 | 4 | 4 | 4 | 5 | 4 | 3 | 4 | 4 | 4 |
| 4 | 5 | 4 | 4 | 5 | 4 | 4 | 5 | 4 | 5 |
| 5 | 4 | 4 | 4 | 4 | 4 | 4 | 5 | 4 | 4 |
| 6 | 4 | 4 | 4 | 4 | 4 | 4 | 5 | 4 | 4 |
| 7 | 4 | 4 | 4 | 4 | 4 | 4 | 5 | 4 | 4 |
| 8 | 4 | 4 | 4 | 4 | 4 | 4 | 5 | 4 | 4 |
| 9 | 4 | 4 | 4 | 4 | 4 | 4 | 5 | 4 | 4 |
| 10 | 4 | 4 | 4 | 4 | 4 | 4 | 5 | 4 | 4 |
| 11 | 4 | 4 | 4 | 4 | 4 | 4 | 4 | 4 | 4 |
| 12 | 4 | 4 | 4 | 4 | 4 | 4 | 5 | 4 | 4 |
| 13 | 5 | 4 | 4 | 4 | 4 | 4 | 4 | 4 | 4 |
| 14 | 4 | 4 | 4 | 5 | 4 | 4 | 4 | 4 | 4 |
| 15 | 4 | 4 | 5 | 4 | 3 | 4 | 5 | 4 | 4 |
| 16 | 5 | 5 | 5 | 5 | 4 | 4 | 5 | 5 | 4 |

*Qwen2-72B-Instruct 020 1188 NP*

| forecaster | criterion 1 | criterion 2 | criterion 3 | criterion 4 | criterion 5 | criterion 6 | criterion 7 | criterion 8 | criterion 9 |
|---|---|---|---|---|---|---|---|---|---|
| 1 | 4 | 5 | 5 | 5 | 5 | 5 | 5 | 5 | 5 |
| 2 | 4 | 4 | 5 | 5 | 4 | 5 | 5 | 5 | 5 |
| 3 | 4 | 5 | 5 | 5 | 5 | 4 | 5 | 5 | 5 |
| 4 | 4 | 4 | 5 | 5 | 5 | 5 | 5 | 5 | 5 |
| 5 | 4 | 5 | 5 | 5 | 5 | 4 | 5 | 5 | 5 |
| 6 | 4 | 4 | 5 | 5 | 5 | 5 | 5 | 5 | 5 |
| 7 | 4 | 4 | 5 | 5 | 5 | 5 | 5 | 5 | 5 |
| 8 | 4 | 4 | 5 | 5 | 5 | 5 | 5 | 5 | 5 |
| 9 | 4 | 4 | 5 | 5 | 5 | 5 | 5 | 5 | 5 |
| 10 | 4 | 4 | 5 | 5 | 5 | 5 | 5 | 5 | 5 |
| 11 | 4 | 4 | 5 | 5 | 5 | 5 | 5 | 5 | 5 |
| 12 | 4 | 4 | 5 | 5 | 5 | 5 | 5 | 5 | 5 |
| 13 | 4 | 4 | 5 | 5 | 5 | 5 | 5 | 5 | 5 |
| 14 | 4 | 4 | 5 | 5 | 5 | 5 | 5 | 5 | 5 |
| 15 | 4 | 4 | 5 | 5 | 5 | 5 | 5 | 5 | 5 |
| 16 | 4 | 4 | 5 | 5 | 5 | 5 | 5 | 5 | 5 |

*DeepSeek-Coder-V2-Instruct 021 1179 NP*

| forecaster | criterion 1 | criterion 2 | criterion 3 | criterion 4 | criterion 5 | criterion 6 | criterion 7 | criterion 8 | criterion 9 |
|---|---|---|---|---|---|---|---|---|---|
| 1 | 4 | 5 | 4 | 5 | 4 | 5 | 4 | 5 | 5 |
| 2 | 4 | 4 | 5 | 4 | 4 | 4 | 4 | 5 | 4 |
| 3 | 4 | 4 | 5 | 3 | 4 | 4 | 4 | 4 | 4 |
| 4 | 5 | 4 | 5 | 4 | 4 | 5 | 5 | 5 | 5 |
| 5 | 4 | 4 | 5 | 4 | 4 | 4 | 4 | 4 | 4 |
| 6 | 4 | 4 | 5 | 4 | 4 | 4 | 4 | 4 | 4 |
| 7 | 4 | 4 | 5 | 4 | 4 | 5 | 4 | 4 | 4 |
| 8 | 4 | 4 | 5 | 4 | 4 | 4 | 4 | 4 | 4 |
| 9 | 4 | 4 | 5 | 4 | 4 | 4 | 4 | 4 | 4 |
| 10 | 4 | 4 | 5 | 4 | 4 | 4 | 4 | 4 | 4 |
| 11 | 4 | 4 | 5 | 4 | 4 | 4 | 4 | 4 | 4 |
| 12 | 4 | 4 | 5 | 4 | 4 | 4 | 4 | 4 | 4 |
| 13 | 4 | 4 | 5 | 4 | 4 | 4 | 4 | 4 | 4 |
| 14 | 4 | 4 | 5 | 4 | 4 | 4 | 4 | 4 | 4 |
| 15 | 4 | 4 | 5 | 4 | 4 | 4 | 4 | 4 | 4 |
| 16 | 5 | 4 | 5 | 4 | 4 | 5 | 5 | 5 | 5 |

*Mistral-Large-2402 029 1157 PP*

| forecaster | criterion 1 | criterion 2 | criterion 3 | criterion 4 | criterion 5 | criterion 6 | criterion 7 | criterion 8 | criterion 9 |
|---|---|---|---|---|---|---|---|---|---|
| 1 | 5 | 4 | 5 | 4 | 5 | 4 | 4 | 5 | 5 |
| 2 | 5 | 4 | 4 | 5 | 5 | 4 | 5 | 4 | 4 |
| 3 | 5 | 4 | 4 | 4 | 4 | 5 | 4 | 5 | 4 |
| 4 | 5 | 3 | 4 | 4 | 4 | 4 | 4 | 4 | 4 |
| 5 | 4 | 4 | 4 | 4 | 4 | 4 | 3 | 4 | 4 |
| 6 | 5 | 4 | 5 | 5 | 5 | 4 | 4 | 5 | 5 |
| 7 | 5 | 4 | 5 | 4 | 4 | 3 | 5 | 4 | 4 |
| 8 | 5 | 4 | 4 | 4 | 4 | 4 | 4 | 4 | 4 |
| 9 | 5 | 4 | 5 | 5 | 4 | 4 | 4 | 5 | 5 |
| 10 | 5 | 4 | 5 | 4 | 5 | 4 | 4 | 5 | 5 |
| 11 | 5 | 4 | 4 | 4 | 4 | 4 | 4 | 5 | 5 |
| 12 | 5 | 4 | 4 | 4 | 4 | 4 | 4 | 4 | 5 |
| 13 | 4 | 4 | 4 | 4 | 4 | 4 | 4 | 4 | 4 |
| 14 | 4 | 3 | 4 | 4 | 3 | 3 | 4 | 4 | 4 |
| 15 | 5 | 4 | 5 | 5 | 4 | 4 | 5 | 5 | 5 |
| 16 | 5 | 4 | 5 | 5 | 5 | 4 | 5 | 5 | 5 |

*Mixtral-8x22b-Instruct-v0.1 035 1146 NP*

| forecaster | criterion 1 | criterion 2 | criterion 3 | criterion 4 | criterion 5 | criterion 6 | criterion 7 | criterion 8 | criterion 9 |
|---|---|---|---|---|---|---|---|---|---|
| 1 | 5 | 5 | 5 | 4 | 4 | 4 | 4 | 5 | 4 |
| 2 | 5 | 5 | 5 | 4 | 4 | 4 | 4 | 5 | 4 |
| 3 | 5 | 5 | 5 | 4 | 4 | 4 | 4 | 5 | 4 |
| 4 | 5 | 5 | 5 | 5 | 4 | 4 | 5 | 5 | 5 |
| 5 | 4 | 5 | 4 | 4 | 4 | 5 | 5 | 4 | 4 |
| 6 | 4 | 5 | 4 | 4 | 4 | 5 | 5 | 4 | 4 |
| 7 | 5 | 5 | 5 | 5 | 5 | 5 | 5 | 5 | 5 |
| 8 | 5 | 5 | 5 | 5 | 4 | 5 | 5 | 4 | 4 |
| 9 | 4 | 5 | 5 | 4 | 4 | 5 | 4 | 4 | 4 |
| 10 | 5 | 5 | 5 | 5 | 4 | 5 | 5 | 5 | 5 |
| 11 | 4 | 5 | 5 | 5 | 4 | 5 | 5 | 4 | 4 |
| 12 | 5 | 5 | 5 | 5 | 4 | 5 | 5 | 5 | 4 |
| 13 | 5 | 5 | 5 | 5 | 4 | 5 | 5 | 5 | 5 |
| 14 | 5 | 5 | 5 | 4 | 4 | 5 | 5 | 4 | 4 |
| 15 | 5 | 5 | 5 | 5 | 5 | 5 | 5 | 5 | 5 |
| 16 | 5 | 5 | 5 | 4 | 5 | 5 | 5 | 4 | 5 |

*Phi-3-Medium-4k-Instruct NP 041 1123 NP*

| forecaster | criterion 1 | criterion 2 | criterion 3 | criterion 4 | criterion 5 | criterion 6 | criterion 7 | criterion 8 | criterion 9 |
|---|---|---|---|---|---|---|---|---|---|
| 1 | 5 | 5 | 5 | 4 | 5 | 4 | 5 | 4 | 4 |
| 2 | 4 | 4 | 5 | 4 | 4 | 5 | 5 | 5 | 5 |
| 3 | 4 | 4 | 5 | 3 | 4 | 3 | 5 | 5 | 4 |
| 4 | 4 | 3 | 5 | 5 | 3 | 4 | 4 | 4 | 4 |
| 5 | 5 | 5 | 5 | 5 | 5 | 5 | 5 | 5 | 5 |
| 6 | 5 | 3 | 4 | 4 | 3 | 4 | 4 | 4 | 4 |
| 7 | 5 | 4 | 4 | 4 | 5 | 5 | 5 | 5 | 4 |
| 8 | 4 | 3 | 5 | 3 | 3 | 2 | 4 | 4 | 3 |
| 9 | 4 | 4 | 3 | 4 | 4 | 3 | 4 | 4 | 3 |
| 10 | 5 | 4 | 5 | 4 | 4 | 5 | 4 | 4 | 4 |
| 11 | 4 | 4 | 5 | 4 | 4 | 4 | 5 | 4 | 4 |
| 12 | 5 | 4 | 5 | 4 | 3 | 3 | 4 | 4 | 4 |
| 13 | 5 | 5 | 5 | 4 | 4 | 5 | 3 | 4 | 4 |
| 14 | 5 | 4 | 5 | 4 | 5 | 4 | 4 | 5 | 5 |
| 15 | 5 | 4 | 4 | 5 | 4 | 5 | 5 | 4 | 4 |
| 16 | 5 | 4 | 5 | 3 | 4 | 4 | 4 | 4 | 4 |

*DBRX-Instruct-Preview 052 1103 NP*

| forecaster | criterion 1 | criterion 2 | criterion 3 | criterion 4 | criterion 5 | criterion 6 | criterion 7 | criterion 8 | criterion 9 |
|---|---|---|---|---|---|---|---|---|---|
| 1 | 4 | 4 | 5 | 4 | 5 | 5 | 5 | 4 | 5 |
| 2 | 4 | 4 | 5 | 4 | 5 | 5 | 4 | 5 | 4 |
| 3 | 4 | 4 | 5 | 4 | 4 | 5 | 5 | 4 | 4 |
| 4 | 5 | 4 | 5 | 5 | 5 | 5 | 5 | 4 | 5 |
| 5 | 5 | 5 | 5 | 5 | 5 | 5 | 4 | 4 | 5 |
| 6 | 5 | 5 | 5 | 5 | 5 | 5 | 5 | 4 | 5 |
| 7 | 5 | 5 | 5 | 5 | 5 | 5 | 5 | 5 | 5 |
| 8 | 5 | 4 | 5 | 5 | 5 | 5 | 5 | 5 | 5 |
| 9 | 5 | 5 | 5 | 5 | 5 | 5 | 4 | 5 | 4 |
| 10 | 5 | 5 | 5 | 5 | 5 | 5 | 5 | 5 | 5 |
| 11 | 5 | 5 | 5 | 5 | 5 | 5 | 5 | 5 | 4 |
| 12 | 5 | 5 | 5 | 5 | 5 | 5 | 5 | 4 | 5 |
| 13 | 5 | 5 | 5 | 5 | 5 | 5 | 5 | 5 | 4 |
| 14 | 5 | 4 | 5 | 5 | 5 | 5 | 5 | 5 | 5 |
| 15 | 5 | 5 | 5 | 5 | 5 | 5 | 5 | 5 | 5 |
| 16 | 5 | 5 | 5 | 5 | 5 | 5 | 4 | 4 | 5 |

*pplx-70b-online 065 1078 PP*

| forecaster | criterion 1 | criterion 2 | criterion 3 | criterion 4 | criterion 5 | criterion 6 | criterion 7 | criterion 8 | criterion 9 |
|---|---|---|---|---|---|---|---|---|---|
| 1 | 5 | 4 | 5 | 4 | 4 | 3 | 5 | 5 | 4 |
| 2 | 5 | 4 | 5 | 5 | 4 | 3 | 5 | 5 | 4 |
| 3 | 5 | 4 | 5 | 4 | 3 | 3 | 5 | 5 | 4 |
| 4 | 5 | 4 | 5 | 5 | 4 | 4 | 5 | 5 | 4 |
| 5 | 5 | 5 | 5 | 5 | 4 | 4 | 5 | 5 | 4 |
| 6 | 5 | 4 | 5 | 5 | 4 | 4 | 5 | 4 | 4 |
| 7 | 5 | 5 | 5 | 5 | 4 | 5 | 5 | 5 | 5 |
| 8 | 5 | 4 | 5 | 5 | 4 | 4 | 5 | 5 | 4 |
| 9 | 5 | 5 | 5 | 5 | 5 | 5 | 5 | 5 | 4 |
| 10 | 5 | 5 | 5 | 5 | 5 | 5 | 5 | 5 | 5 |
| 11 | 5 | 5 | 5 | 5 | 5 | 5 | 5 | 5 | 5 |
| 12 | 5 | 5 | 5 | 5 | 5 | 5 | 5 | 5 | 5 |
| 13 | 5 | 5 | 5 | 5 | 5 | 5 | 5 | 4 | 5 |
| 14 | 5 | 4 | 5 | 5 | 5 | 4 | 5 | 5 | 5 |
| 15 | 5 | 5 | 5 | 5 | 4 | 5 | 5 | 5 | 5 |
| 16 | 5 | 5 | 5 | 5 | 5 | 5 | 5 | 5 | 5 |

**Table 5.** Evaluation score given to the i-th forecaster for criterion k, averaged across all raters. Cell (j,k) in this table is computed as $S_i^{(Ck)} = \frac{1}{16}\sum_{j=1}^{16} S_{ij}^{(Ck)} \in R^{16\times 9}$ .

| **Forecaster** | **C1** | **C2** | **C3** | **C4** | **C5** | **C6** | **C7** | **C8** | **C9** | ***Avg F*** |
|---|---|---|---|---|---|---|---|---|---|---|
| gpt-4o | 4,438 | 4,250 | 4,500 | 3,875 | 4,000 | 3,813 | 4,375 | 4,375 | 4,313 | *4,215* |
| claude-3-5-sonnet | 4,313 | 4,250 | 4,625 | 4,375 | 4,125 | 3,813 | 4,625 | 4,500 | 4,188 | *4,313* |
| gemini-1-5-pro-api | 4,125 | 3,875 | 4,375 | 3,875 | 3,500 | 3,500 | 4,375 | 4,250 | 3,875 | *3,972* |
| Yi-Large-preview | 4,688 | 3,875 | 4,688 | 4,438 | 3,750 | 4,188 | 4,563 | 4,313 | 4,313 | *4,313* |
| Gemma-2-27B | 4,125 | 4,250 | 4,375 | 4,063 | 3,625 | 4,000 | 3,938 | 4,000 | 3,938 | *4,035* |
| GLM-4-0520 | 4,375 | 4,000 | 4,438 | 4,250 | 3,938 | 4,063 | 4,438 | 3,938 | 4,000 | *4,160* |
| Llama-3-70b-Instruct | 4,375 | 4,250 | 4,500 | 4,313 | 4,125 | 4,313 | 4,688 | 4,250 | 4,125 | *4,326* |
| Reka-Core-2 | 4,250 | 3,875 | 4,563 | 4,000 | 3,750 | 3,875 | 4,063 | 3,938 | 3,813 | *4,014* |
| Command-R+ | 4,313 | 4,375 | 4,250 | 4,188 | 4,063 | 4,250 | 4,125 | 4,063 | 3,875 | *4,167* |
| Qwen2-72B-Instruct | 4,625 | 4,250 | 4,750 | 4,250 | 4,250 | 4,313 | 4,500 | 4,125 | 4,313 | *4,375* |
| DeepSeek-Coder-V2 | 4,375 | 3,938 | 4,250 | 4,063 | 4,000 | 4,000 | 4,438 | 4,063 | 3,875 | *4,111* |
| Mistral-Large-2402 | 4,375 | 4,188 | 4,625 | 4,125 | 3,938 | 4,063 | 4,438 | 4,000 | 3,938 | *4,188* |
| Mixtral-8x22b-Instruct | 4,313 | 4,375 | 4,500 | 4,125 | 4,250 | 4,438 | 4,063 | 3,813 | 3,875 | *4,194* |
| Phi-3-Medium-4k-Instruct | 4,250 | 3,813 | 4,438 | 4,188 | 4,125 | 3,938 | 4,438 | 4,188 | 4,063 | *4,160* |
| DBRX-Instruct-Preview | 4,375 | 4,250 | 4,563 | 4,313 | 3,938 | 4,438 | 4,438 | 4,000 | 4,063 | *4,264* |
| pplx-70b-online | 4,875 | 4,000 | 4,938 | 4,313 | 4,375 | 4,438 | 4,750 | 4,438 | 4,438 | *4,507* |
| ***Average C score*** | *4,383* | *4,110* | *4,521* | *4,168* | *3,983* | *4,090* | *4,387* | *4,137* | *4,063* | *4,207* |
| ***Std deviation C score*** | *0.190* | *0.189* | *0.174* | *0.162* | *0.229* | *0.261* | *0.226* | *0.192* | *0.191* | *0,136* |

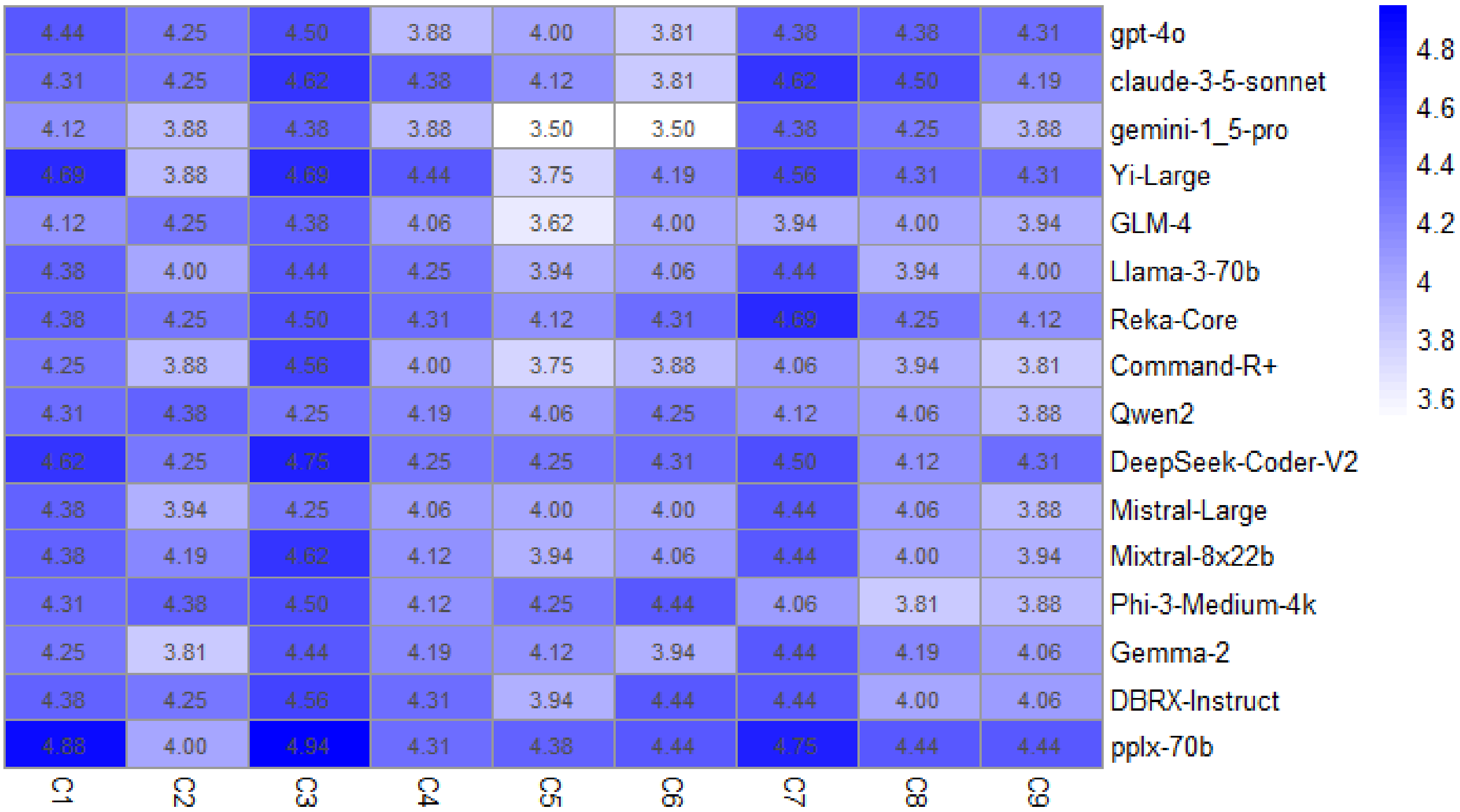

**Figure 2.** Heat map of the gaps between the evaluation score for a single criterium (k-th) and average score of the criterium, normalized to the standard deviation of the criterium: $(S_j^{(Ck)} - S_{av}^{(Ck)})/S_{std}^{(Ck)}$.

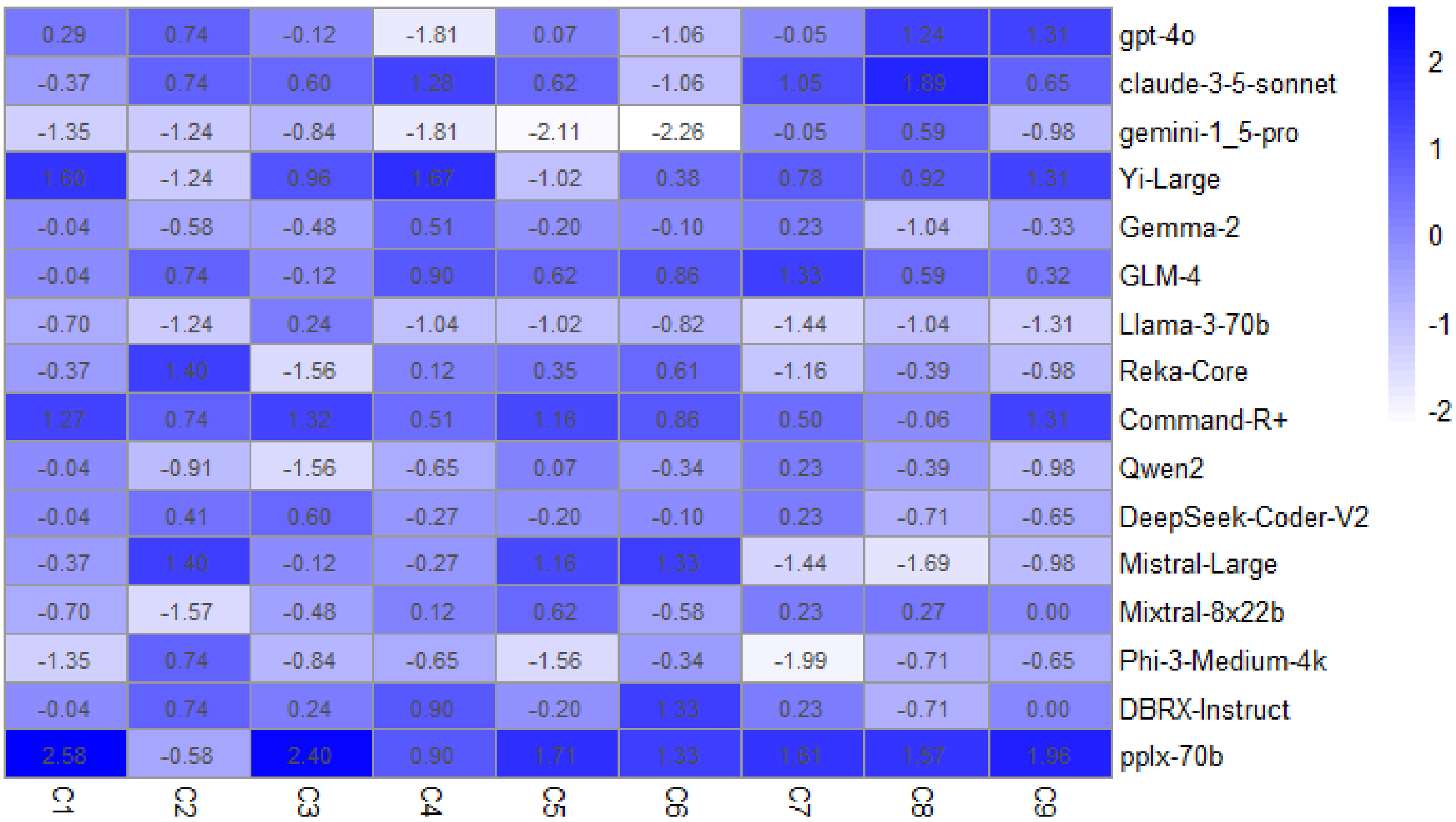

| | *C1* | *C2* | *C3* | *C4* | *C5* | *C6* | *C7* | *C8* | *C9* |
|---|---|---|---|---|---|---|---|---|---|
| *gpt-4o* | *0,289* | *0,741* | *-0,121* | *-1,812* | *0,074* | *-1,063* | *-0,053* | *1,240* | *1,306* |
| *claude-3-5-sonnet* | *-0,368* | *0,741* | *0,598* | *1,280* | *0,620* | *-1,063* | *1,054* | *1,891* | *0,653* |
| gemini-1_5-pro | -1,355 | -1,243 | -0,840 | -1,812 | -2,108 | -2,264 | -0,053 | 0,589 | -0,982 |
| Yi-Large | 1,602 | -1,243 | 0,960 | 1,670 | -1,017 | 0,376 | 0,779 | 0,917 | 1,306 |
| Gemma-2 | -1,355 | 0,741 | -0,840 | -0,649 | -1,563 | -0,345 | -1,988 | -0,714 | -0,653 |
| GLM-4 | -0,042 | -0,582 | -0,477 | 0,507 | -0,196 | -0,104 | 0,226 | -1,037 | -0,329 |
| **Llama-3-70b** | -0,042 | 0,741 | -0,121 | 0,897 | 0,620 | 0,856 | **1,333** | 0,589 | 0,324 |
| Reka-Core | -0,699 | -1,243 | 0,242 | -1,039 | -1,017 | -0,825 | -1,435 | -1,037 | -1,306 |
| Command-R+ | -0,368 | 1,402 | -1,558 | 0,124 | 0,349 | 0,614 | -1,160 | -0,386 | -0,982 |
| **Qwen2** | **1,271** | 0,741 | 1,317 | 0,507 | **1,165** | 0,856 | 0,500 | -0,063 | 1,306 |
| DeepSeek-Coder-V2 | -0,042 | -0,910 | -1,558 | -0,649 | 0,074 | -0,345 | 0,226 | -0,386 | -0,982 |
| Mistral-Large | -0,042 | 0,413 | 0,598 | -0,266 | -0,196 | -0,104 | 0,226 | -0,714 | -0,653 |
| Mixtral-8x22b | -0,368 | 1,402 | -0,121 | -0,266 | 1,165 | 1,335 | -1,435 | -1,688 | -0,982 |
| Phi-3-Medium-4k | -0,699 | -1,571 | -0,477 | 0,124 | 0,620 | -0,583 | 0,226 | 0,266 | 0,000 |
| DBRX-Instruct | -0,042 | 0,741 | 0,242 | 0,897 | -0,196 | 1,335 | 0,226 | -0,714 | 0,000 |
| **pplx-70b** | **2,584** | -0,582 | 2,398 | 0,897 | **1,711** | **1,335** | **1,607** | **1,568** | **1,958** |

**11.3** Average self-assessments score vs average of evaluations received from others.

**Table A.3**. Average self-assessments score vs average of evaluations received from others.

| | SES | HES | SEI |
|---|---|---|---|
| Residuals | Min 1Q Median 3Q Max<br>-1.26733 -0.20091 0.03639 0.32197 0.88886 | Min 1Q Median 3Q Max<br>-0.22100 -0.08848 -0.03593 0.09270 0.23330 | Min 1Q Median 3Q Max<br>-0.287395 -0.039277 0.005258 0.085184 0.183022 |
| Coefficients<br><br>Signif. codes: 0 '***' 0.001 '**' 0.01 '*' 0.05 '.' 0.1 ' ' 1 | Estimate Std. Error t value Pr(>\|t\|)<br>(Intercept) 12.879214 3.199777 4.025 0.00125 **<br>arenaScore -0.007239 | Estimate Std. Error t value Pr(>\|t\|)<br>(Intercept) 4.6504813 0.6548476 7.102 5.32e-06 ***<br>arenaScore -0.0003746 | Estimate Std. Error t value Pr(>\|t\|)<br>(Intercept) 2.962641 0.728531 4.067 0.00116 **<br>arenaScore -0.001638 |
| | 0.002684 -2.698 0.01734 * | 0.0005492 -0.682 0.506 | 0.000611 -2.681 0.01790 * |
| | Residual standard error: 0.6167 on 14 DF<br>Multiple R-squared: 0.342,<br>Adjusted R-squared: 0.295<br>F-statistic: 7.277 on 1 and 14 DF,<br>p-value: 0.0173 | Residual standard error: 0.1262 on 14 DF<br>Multiple R-squared: 0.03216<br>Adjusted R-squared: -0.03697<br>F-statistic: 0.465 on 1 and 14 DF<br>p-value: 0.506 | Residual standard error: 0.1404 on 14 DF<br>Multiple R-squared: 0.3393<br>Adjusted R-squared: 0.2921<br>F-statistic: 7.19 on 1 and 14 DF<br>p-value: 0.0179 |

**11.4 Conversion of likelihood estimate to Expert score**

The expert-derived scores are computed according to the ratio between the LLM and the expert estimates of the likelihood of the AGI event. Being f_i the mean likelihood estimate of i-th LLM's, we have:

```
agiLH <- c(45,5.8,12.5,38,5,8,15,3,15,15,5,12,15,6.3,3.5,47.6)

fc <- c(0,0.4,0,0,0.4,0.39,0,0.39,0,0,0.4,0,0,0.39,0.4,0)

grace <- 10

ratio <- agiLH / grace

adj1 <- ratio[12]

adj2 <- ratio[16]-ratio[12]

delta <- abs((ratio - adj1)/adj2)

delta[12] <- 0.01

exp_sco <- round(5 - (delta*4),2)+fc
```

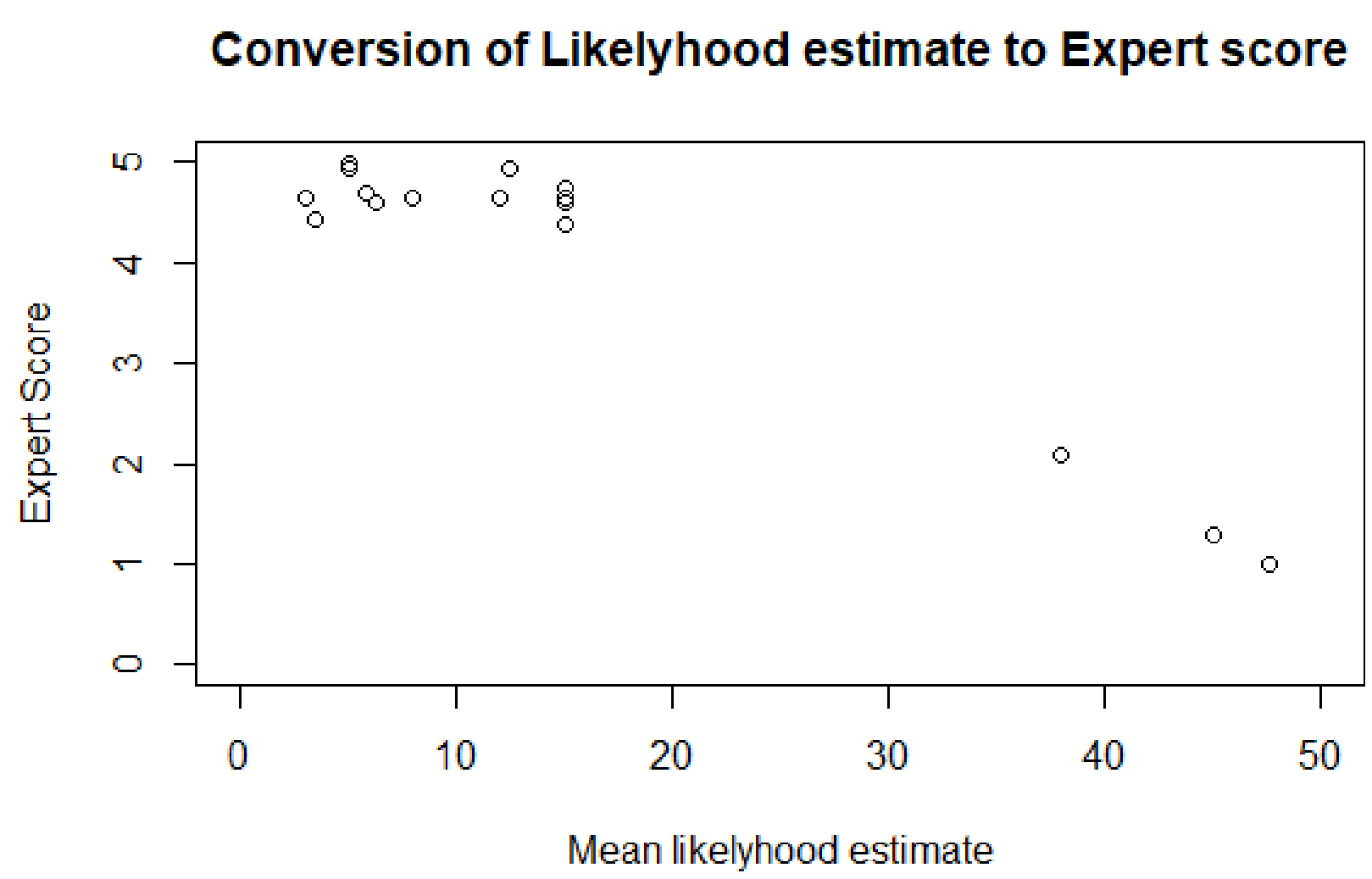

**11.5 Optimization codes**

***L-BFGS-B - alpha 1, beta 73***

```
objective_function <- function(x, S, newHJS, rankNewHJS, alpha, beta) {
  term1 <- alpha * sum((newHJS - S %*% x)^2)
  term2 <- beta * normDistKendall(rank(-as.vector(S %*% x), ties.method = "min"), rankNewHJS, "ranking")
  return(term1 + term2)}
```

```
# Parameters
n <- 16
alpha <- 1
beta <- 73
start_point <- rep(1/16, n)  # Punto di partenza
# constrain definition
lower_bounds <- rep(0, n)  # Limite inferiore: tutti gli elementi >= 0
upper_bounds <- rep(1, n)  # Limite superiore: tutti gli elementi <= 1
# Funzione per proiettare il vettore sulla somma unitaria
project_to_simplex <- function(x) {
  return(x / sum(x))}
# Funzione wrapper per l'ottimizzazione
optim_wrapper <- function(x, S, newHJS, rankNewHJS, alpha, beta) {
  x_proj <- project_to_simplex(x)
  return(objective_function(x_proj, S, newHJS, rankNewHJS, alpha, beta))}
# Esecuzione dell'ottimizzazione
result <- optim(
  par = start_point,
  fn = optim_wrapper,
  method = "L-BFGS-B",
  lower = lower_bounds,
  upper = upper_bounds,
  control = list(maxit = 10000),
  S = S,
  newHJS = newHJS,
  rankNewHJS = rankNewHJS,
  alpha = alpha,
  beta = beta)
# Estrazione e proiezione del risultato ottimale
optimal_x <- project_to_simplex(result$par)
#routine di controllo
cat(optimal_x, sep = "\n")
sum(optimal_x)
plot(optimal_x)
S %*% optimal_x
(S %*% optimal_x)/ newHJS
cat(rank(-(S %*% optimal_x), ties.method = "min"), sep = "\n")
distKendall(rank(-(S %*% optimal_x), ties.method = "min"), rankNewHJS, type =  "ranking")
normDistKendall(rank(-(S %*% optimal_x), ties.method = "min"), rankNewHJS, "ranking")
```

```
confronta_ranking_v2(rank(-(S %*% optimal_x)), rankNewHJS)
distKendall(rank(-(S %*% optimal_x), ties.method = "min"), rank_Arena, type =  "ranking")
normDistKendall(rank(-(S %*% optimal_x), ties.method = "min"), rank_Arena, "ranking")
confronta_ranking_v2(rank(-(S %*% optimal_x)), rank_Arena)
t(newHJS - S %*% optimal_x) %*% (newHJS - S %*% optimal_x)*alpha
normDistKendall(rank(-(S %*% optimal_x), ties.method = "min"), rankNewHJS, "ranking")*beta
```

***DEoptim - alpha 1, beta 17***

```
library(DEoptim)
# Definizione della funzione obiettivo con penalizzazione per i vincoli
objective_function <- function(x, S_14, newHJS_14, rankNewHJS_14, alpha, beta, penalty = 1e6) {
  # Calcolo della funzione obiettivo originale
  term1 <- alpha * sum((newHJS_14 - S_14 %*% x)^2)
  term2 <- beta * normDistKendall(rank(-as.vector(S_14 %*% x), ties.method = "min"), rankNewHJS_14, "ranking")
  obj <- term1 + term2
  # Penalizzazione per il vincolo di somma uguale a 1
  penalty_sum <- penalty * (sum(x) - 1)^2
  # Penalizzazione per il vincolo di non negatività non è necessaria
  # poiché DEoptim può gestire direttamente i limiti inferiori e superiori
  return(obj + penalty_sum)}
# Parametri
alpha <- 1
beta <- 17
n <- 14  # dimensione del vettore x
# Definizione dei limiti inferiori e superiori
lower <- rep(0, n)  # Limite inferiore: 0 per tutti gli elementi
upper <- rep(1, n)  # Limite superiore: 1 per tutti gli elementi
# Funzione wrapper per passare i parametri fissi a DEoptim
obj_wrapper <- function(x) {
  objective_function(x, S_14, newHJS_14, rankNewHJS_14, alpha, beta)}
# Ottimizzazione con DEoptim
result <- DEoptim(
  fn = obj_wrapper,
  lower = lower,
  upper = upper,
  control = DEoptim.control(
    NP = 10 * n,  # Dimensione della popolazione
    itermax = 1100,  # Numero massimo di iterazioni
    F = 0.8,  # Fattore di differenziazione
```

```
    CR = 0.9,  # Probabilità di crossover
    strategy = 2,  # Strategia DE/local-to-best/1/bin
    trace = TRUE )) # Per visualizzare il progresso dell'ottimizzazione
# Estrai il miglior risultato
optimal_x <- result$optim$bestmem
#routine di controllo
cat(optimal_x, sep = "\n")
sum(optimal_x)
plot(optimal_x)
S_14 %*% optimal_x
(S_14 %*% optimal_x)/ newHJS_14
cat(rank(-(S_14 %*% optimal_x), ties.method = "min"), sep = "\n")
distKendall(rank(-(S_14 %*% optimal_x), ties.method = "min"), rankNewHJS_14, type =  "ranking")
normDistKendall(rank(-(S_14 %*% optimal_x), ties.method = "min"), rankNewHJS_14, "ranking")
confronta_ranking_v2(rank(-(S_14 %*% optimal_x)), rankNewHJS_14)
distKendall(rank(-(S_14 %*% optimal_x), ties.method = "min"), rank_Arena_14, type =  "ranking")
normDistKendall(rank(-(S_14 %*% optimal_x), ties.method = "min"), rank_Arena_14, "ranking")
confronta_ranking_v2(rank(-(S_14 %*% optimal_x)), rank_Arena_14)
t(newHJS_14 - S_14 %*% optimal_x) %*% (newHJS_14 - S_14 %*% optimal_x)*alpha
normDistKendall(rank(-(S_14 %*% optimal_x), ties.method = "min"), rankNewHJS_14, "ranking")*beta
```

**11.6 Function**

*Normalized distance Kendall*

```
library(Rankcluster)
normDistKendall <- function(vettore1, vettore2, tipo_stringa) {
    # Verifica che i due vettori abbiano lo stesso numero di elementi
    if (length(vettore1) != length(vettore2)) {
        stop("I due vettori devono avere lo stesso numero di elementi.")    }
        # Calcola la distanza di Kendall utilizzando la funzione distKendall con il parametro 'type'
    distanza_kendall <- distKendall(vettore1, vettore2, type = tipo_stringa)
     # Calcola il numero di elementi
    n <- length(vettore1)
    # Calcola la distanza massima di Kendall per n elementi
    distanza_kendall_massima <- n * (n - 1) / 2
     # Normalizza la distanza di Kendall
    distanza_kendall_normalizzata <- distanza_kendall / distanza_kendall_massima
     # Restituisce la distanza normalizzata
    return(distanza_kendall_normalizzata)}
```

*Confronta ranking*

```
confronta_ranking_v2 <- function(vettore1, vettore2) {
  # Controlla che i due vettori abbiano la stessa lunghezza
  if (length(vettore1) != length(vettore2)) {
    stop("I due vettori devono avere la stessa lunghezza")
  }
    # Trova le posizioni dove i valori coincidono
  coincidenze <- vettore1 == vettore2
    # Estrai i valori che coincidono
  valori_coincidenti <- vettore1[coincidenze]
    # Conta il numero di valori coincidenti
  numero_coincidenti <- sum(coincidenze)
    # Restituisci i valori coincidenti e il loro numero
  return(list(valori_coincidenti = valori_coincidenti, numero_coincidenti = numero_coincidenti))}
```